\documentclass[lettersize,journal]{IEEEtran}
\usepackage{amsmath,amsfonts}
\usepackage{algorithmic}
\usepackage{algorithm}
\usepackage{array}
\usepackage[caption=false,font=normalsize,labelfont=sf,textfont=sf]{subfig}
\usepackage{textcomp}
\usepackage{stfloats}
\usepackage{url}
\usepackage{verbatim}
\usepackage{graphicx}
\usepackage{cite}

\newlength{\noname}
\newcommand{\anonymize}[1]{%
    \ifdefined\anon
        \settowidth{\noname}{#1}%
        \rule{\noname}{.6\baselineskip}%
    \else
        #1%
    \fi
}

\begin{document}

\title{Identifying the Key Biomechanical Features of Movement Adaptation during Exoskeleton-Assisted Locomotion}

\author{\anonymize{Peter Seungjune Lee,~\IEEEmembership{Student Member,~IEEE,} and Katja Mombaur,~\IEEEmembership{Senior Member,~IEEE}}
\thanks{This work involved human subjects in its research. Approval of all ethical and experimental procedures and protocols was granted by the \anonymize{Karlsruhe Institute of Technology (KIT)} Ethics Committee under Application No. \anonymize{A2024-007}. This work was made possible thanks to the support of the chair by \anonymize{the Hector Foundation II.}}
\thanks{\anonymize{Peter Seungjune Lee and Katja Mombaur} are with the \anonymize{KIT BioRobotics} \anonymize{Lab, Optimization and Biomechanics for Human-Centred Robotics, Institute} \anonymize{for Anthropomatics and Robotics, Karlsruhe Institute of Technology, 76131} \anonymize{Karlsruhe, Germany (email: seungjune.lee@kit.edu; katja.mombaur@kit.edu)}}
\thanks{\anonymize{Katja Mombaur} is also with the \anonymize{Human-Centred Robotics and Machine} \anonymize{Intelligence, department of systems design engineering, University of Waterloo,} \anonymize{Waterloo, N2L3G1, Canada (email: katja.mombaur@uwaterloo.ca)}}
}
\markboth{IEEE TRANSACTIONS ON MEDICAL ROBOTICS AND BIONICS,~Vol.~X, No.~X, X~2025}%
{Shell \MakeLowercase{\textit{et al.}}: A Sample Article Using IEEEtran.cls for IEEE Journals}

\IEEEpubid{0000--0000/00\$00.00~\copyright~2021 IEEE}

\maketitle

\begin{abstract}
The understanding of natural human adaptation during exoskeleton-assisted locomotion - particularly individual differences in adaptation behaviors and temporal progression - remains limited. In this work, we investigate temporal evolution of biomechanical variables to uncover participant-specific adaptation strategies across different exoskeleton-assisted locomotion scenarios. Nine healthy participants performed treadmill walking under three conditions: without an exoskeleton, with exoskeleton active ankle assistance, and with exoskeleton zero-torque. Lower limb kinematics, inter-joint coordination, and metabolic cost of transport (MCoT) were analyzed at both the group and individual levels. Results indicate that adaptation is gradual and highly individualized, with substantial variability in convergence timing and movement patterns across participants. Kinematic adaptation occurred asynchronously across lower limb, with larger fluctuations during the swing phase. Metabolic responses were heterogeneous and often non-convergent, highlighting the limitations of steady-state assumptions commonly adopted in the literature. These findings emphasize the importance of individual-level, temporal evolution analyses for understanding adaptation dynamics in exoskeleton use.

\end{abstract}


\section{Introduction}
\IEEEPARstart{U}{nder} the overarching goal of providing physical assistance to help rehabilitation patients regain mobility independence, numerous active lower limb exoskeleton designs and control strategies have been developed over the years. Despite extensive laboratory experiments and clinical trials demonstrating the efficacy of various lower limb exoskeleton designs \cite{exoReviewClinical}, the detailed comprehension of physical human-exoskeleton interaction (pHEI) remains limited \cite{exoReviewPHRI}. In particular, how natural human adaptation occurs during exoskeleton-assisted locomotion, the underlying objective functions driving this familiarization process, and where this adaptation behavior converge toward are not yet well understood \cite{giorgos, Kirchner, exoReviewIntention}.

Humans possess remarkable adaptive intelligence as part of innate intuition, enabling them to naturally adjust motor strategies to changing environments and varying conditions. This intriguing adaptation process can occur subconsciously in response to external perturbation or assistance, even during complex interactions such as pHEI \cite{subconsciousAdaptation}. Specifically, the exoskeleton imparts predefined trajectories or torque profiles to provide movement assistance during exoskeleton-assisted locomotion. Throughout this process, the human body is subjected to external forces and moments transmitted through the human-exoskeleton interface, to which the users adapt to over time. Recent studies have demonstrated that this motor adaptation to exoskeleton assistance is a complex, time-dependent process involving continuous exploration and learning across extended periods of exposure \cite{poggensee2021,panizzolo2019}. During this adaptation phase, the nervous system actively explores coordination strategies to balance mechanical assistance against altered muscle activation and energetic demands \cite{hybart2023,abram2022}. 

Aside from exoskeleton assistive loads, motion constraints and kinematic misalignment inherent to the specific exoskeleton design further contribute to the complexity of pHEI adaptation \cite{pHEIComplexity,pHEIIssues}. In addition, users may experience momentary mismatches between the prescribed assistance profiles and their intended motion, resulting individualized adaptation responses during exoskeleton use \cite{h2tPaper,pHEIPortrait}. Understanding these individual adaptation responses can provide insights regarding the physical implications of exoskeleton use from the individual user's perspective and hence strategies for better personalization. However, a considerable number of existing works evaluate novel exoskeleton designs and control strategies from a statistical point of view, emphasizing group-level trends rather than focusing on individual biomechanical responses and individual adaptation strategies \cite{conorWalshMetabolic}. 

\IEEEpubidadjcol

While statistical inferences provide useful information regarding the general trend of the control group, they obscure the inter-participant differences and overlook unique adaptation strategies\cite{individualDiff}. For example, changes in joint kinematics is among the most commonly reported performance indicators when assessing exoskeleton designs, due to the ease of measurement from robot sensors and motion capture systems \cite{katjaReviewPaper}. Although reporting changes in joint trajectories, range of motion (ROM), and spatiotemporal parameters is valuable, these variables are typically represented as an average across participants and often reported using means, standard deviations, and p-values \cite{conorWalshSoftExosuit,avgParticipants}. Even among the works considering the inter-participant variabilities \cite{interSubject}, the changes in biomechanical variables are often reported as end-to-end state comparisons, such as averaging the final minutes of metabolic cost or kinematic responses \cite{conorWalshSoftExosuit, 2min, 25min}. This approach, though straightforward to adopt, inherently considers only the steady-state behavior and neglects the temporal aspects of adaptation dynamics, which provide critical insight into individual adaptation processes throughout the exoskeleton use.

\begin{figure*}[t]
  \centering 
  \includegraphics[width=1.0\linewidth]{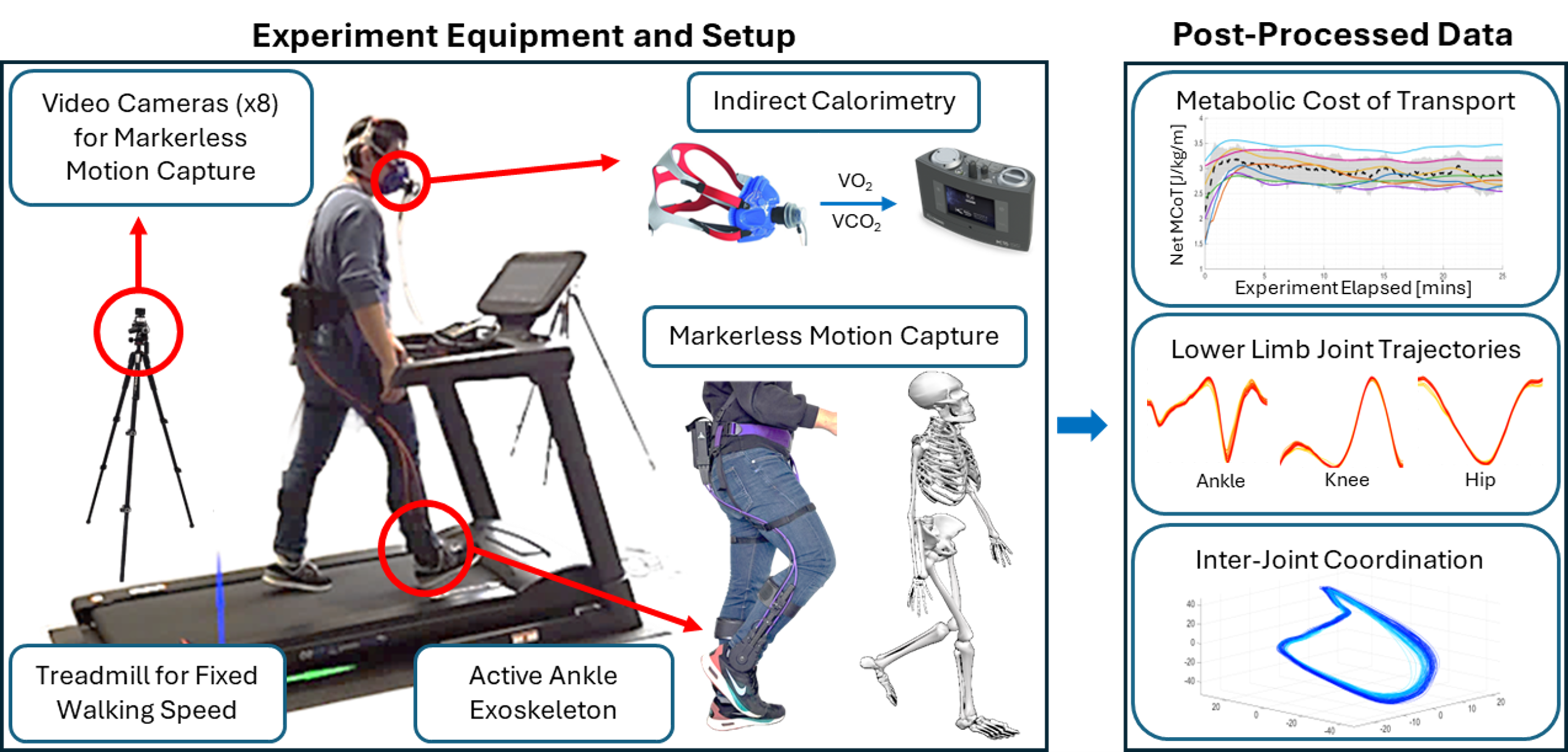}
  \caption{Overview of the experimental setup and post-processed outputs in the human-centered investigation of adaptation during exoskeleton-assisted locomotion.}
  \label{fig:fig1}
\end{figure*}

In this work, we investigate the individual short-term motor adaptation strategies by analyzing the temporal evolution of biomechanical variables during exoskeleton-assisted walking. Specifically, we evaluate how joint-level kinematic variability and inter-joint coordination evolve under bidirectional ankle torque assistance (plantar flexion and dorsiflexion) at participant-specific preferred treadmill walking speeds. The central contribution of the presented work is the metric framework used to quantify short-term motor adaptation, such as joint-level inter-trajectory convergence metrics and the linear dependency analysis derived from elevation angle cyclograms to track the continuous evolution of inter-joint coordination during exoskeleton-assisted locomotion. Along with metabolic cost response monitoring and the self-reported perception of adaptation, the framework is aimed at deducing complex motor adaptation into motor patterns adopted across participants. We hypothesize that: (1) the early phase of motor adaptation to exoskeleton-assisted locomotion induces heterogeneous, transient responses across individuals, (2) that kinematic variance is gait-phase dependent, and (3) that active assistance transiently disrupts the natural lower limb inter-joint coordination patterns.

The remainder of the paper is organized as follows. Section II details the data collection equipment, experimental tasks, and post-processing methods. Section III presents the qualitative and quantitative results for lower limb kinematics and metabolic cost responses. Section IV provides an in-depth discussion of the findings, with specific emphasis on subject specific kinematic adaptation strategies, inter-joint coordination, and methodological considerations for studying adaptation processes in exoskeleton-assisted locomotion. Finally, Section V concludes the paper.

\section{Materials and Methods}
Considering that every individual walks differently and may adopt unique adaptation strategies during the intuitive neuromotor adaptation process to exoskeleton assistance, we conducted a study to observe how the biophysical variables change throughout the different exoskeleton-assisted locomotion scenarios. 
The following sections outline the study details and rationales, including the choice of equipment and the metrics adopted in the analysis of the collected motion capture data.

\subsection{Study Equipment}
\subsubsection{Active Ankle Exoskeleton}
The overview of the equipment used in this study is as shown in Fig. \ref{fig:fig1}. The exoskeleton used is a cable-driven active ankle soft exoskeleton (Spark, Biomotum Inc), which showcased promising results for improving the metabolic responses among children and older adults with neurological disorders \cite{sparkOG, sparkmore}. The system consists of a waist pack housing the controller electronics, battery, and two motors, each driving Bowden cables connected to the revolute joint at the ankle. The exoskeleton can provide bidirectional assistance in both plantar flexion and dorsiflexion with predefined torque profiles during gait. The magnitude of torque profiles are proportionally scaled based on the forces at the ball of the foot measured by the force sensor on the footplate \cite{spark}.

\begin{figure*}
    \centering
    \includegraphics[width=1.0\linewidth]{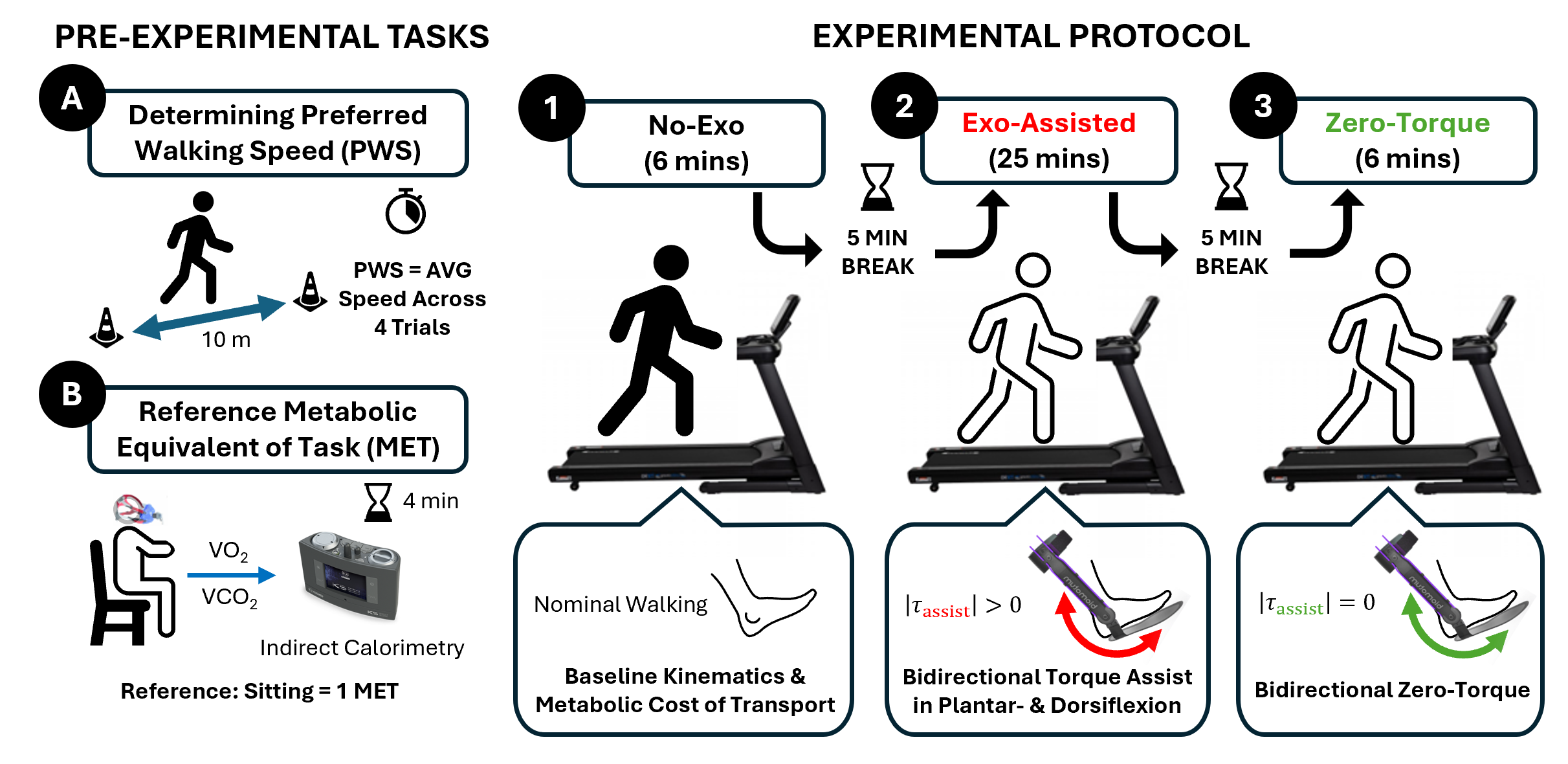}
    \caption{A schematic illustrating the pre-experimental tasks and the experimental protocols. Note that in Exo-Assisted walking task, the exoskeleton is providing bidirectional torque assistance throughout the gait, in both plantar flexion and dorsiflexion. During Exo Zero-Torque walking task, the exoskeleton is set to provide zero torque at the ankle, minimizing the impedance to human ankle movement.}
    \label{fig:protocol}
\end{figure*}

\subsubsection{Motion Capture System}
The joint kinematics data were attained via the markerless motion capture system (Theia3D, Theia Markerless). As opposed to the standard marker-based motion capture systems requiring infrared cameras and manual placement of reflective markers, markerless motion capture only requires the video footage of the experiments recorded with regular video cameras. The recorded video footage are fed into the proprietary post-processing software, in which it outputs subject-specific kinematic model and joint trajectory estimations. The validity of this markerless motion capture system has previously been verified to be highly accurate, especially for the lower limb joint motions in the sagittal plane in gait and running \cite{theiaValidLowerLimbSagittal,theiaValidWalkingRunning}.

The capture space was configured using 8 digital cameras synchronized via the camera control box (RX0 ii and CCB-WD1, Sony). The calibration was done such that all segment coordinate systems are presented in the positive axes of the global lab coordinate system. Although the optimal settings may be dependent on the specific lab environment, the videos were captured at 120 frames per second (fps) with shutter speed of 1/400 s, which provided minimal blur and sufficient lighting for optimal calibration root mean squared error (RMSE). With this setup, the participants' kinematics data were seamlessly generated without occlusion, even when exoskeleton was donned onto the participants. 

\subsubsection{Metabolic Cost}
The metabolic response data were recorded using the portable indirect calorimetry monitoring system (K5, Cosmed). 

\subsection{Experiment}
\subsubsection{Participants}
A total of 9 healthy young adults (6 female and 3 male participants) with no ongoing or history of musculoskeletal injuries or mobility impairments participated in this study. Prior to the participation of the experiment, mobility-related questionnaire was filled out for additional health screening. As this is a study designed for observing the course of the changes in biomechanical variables during the early stages of the natural motor adaptation process, applicants with prior experience with the exoskeleton used in this study were excluded from participation. The participant pool is as summarized in Table \ref{table:table1}. 
All participants were given transparent details of the study and have provided written consent prior to the commencement of the experiment. The experimental protocol was reviewed and approved by the \anonymize{Karlsruhe Institute of Technology} ethics committee \anonymize{(A2024-007)}. The data protection methods including data storage, pseudonymization, and processing were carried out in accordance with the European Union General Data Protection Regulation (GDPR).

\renewcommand{\arraystretch}{1.5}
\begin{table}[t]
    \centering
    \caption{Summary of participant characteristics and preferred walking speeds obtained in the pre-experimental task}
    \label{table:table1}
    \begin{tabular}{|c|c|c|c|}
        \hline
        \textbf{Age} &
        \textbf{Height (cm)} &
        \textbf{Weight (kg)} &
        \textbf{Walking Speed (m/s)} \\
        \hline 
        25.7 $\pm$ 3.50 & 171.6 $\pm$ 9.99 & 69.0 $\pm$ 13.03 & 1.14 $\pm$ 0.16 \\
        \hline
    \end{tabular}
\end{table}

\subsubsection{Pre-Experimental Tasks}
A schematic illustrating the pre-experimental and the experimental tasks are as shown in Fig. \ref{fig:protocol}. Before proceeding to the experimental tasks, the individual preferred walking speeds were first determined for each participant. Specifically, each participant was asked to walk a 10-meter interval at their own pace after being given the prompt: \textit{Imagine you are going for a walk. For example, pretend you are walking by the \anonymize{Rhein river} for about 30 minutes on a sunny day}. The averaged speed across four trials was chosen as the preferred walking speed, which was then verified with the participant on the treadmill. Placing 1-meter buffers on each end of the 10-meter markers ensured the participants had sufficient time to accelerate to and decelerate from their natural walking speeds. As opposed to making the participants walk at a fixed speed predefined for the experiment, which is usually set at relatively high walking speeds without reasonable justifications and often not considering the different participant demographics \cite{speed}, this method takes into the account the subjective speed preferences and the different perception of natural walking efforts.

Lastly, metabolic measurement system was fitted on the participant and the resting metabolic rate (RMR) was collected during 4 minutes of sitting, establishing the reference Metabolic Equivalent of Task (MET) \cite{MET}. To ensure reliable metabolic responses, participants were instructed to avoid excessive exercise prior to the experiment and refrain from eating, especially from consuming caffeine, for at least two hours beforehand \cite{metabolicNoFood}.

\begin{figure*}[t]
  \centering 
  \includegraphics[width=1.0\linewidth]{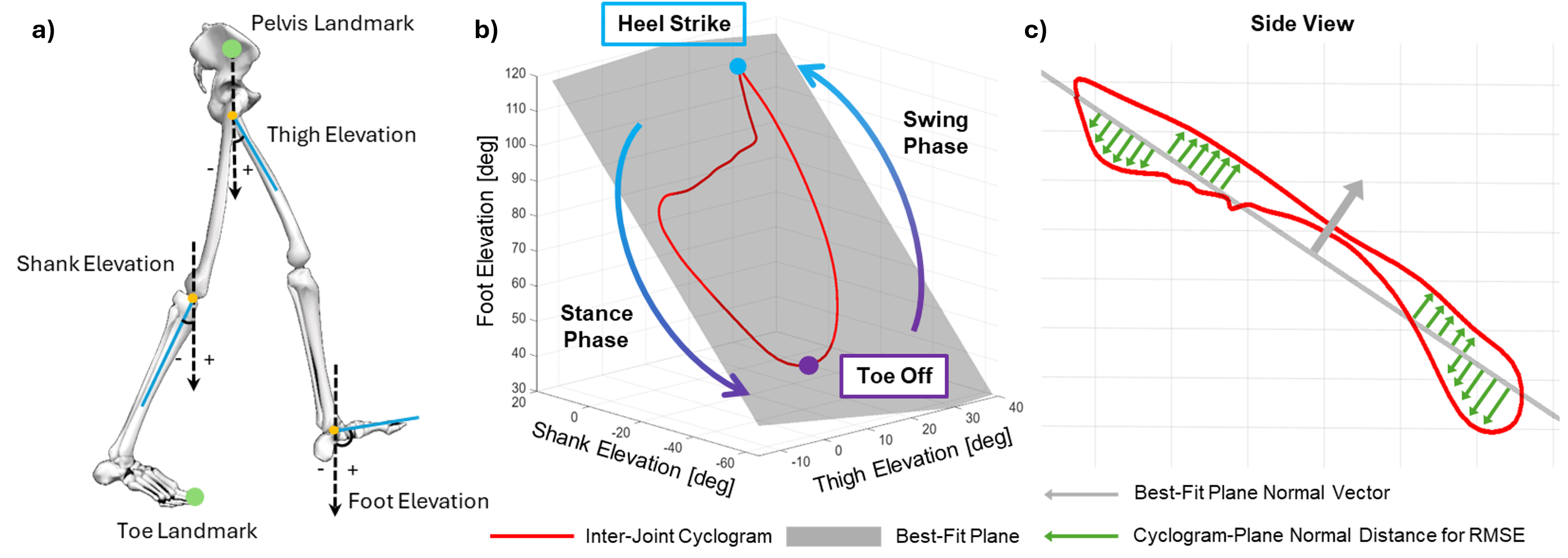}
  \caption{Illustration of a) the definition of segment elevation angles in sagittal plane for inter-joint coordination analysis between thigh, shank, and foot, computed relative to the gravity vector. The locations of the pelvis and the toe landmarks displayed in green are used for kinematics-based gait segmentation. b) Representative three-dimensional inter-joint cyclogram of a single gait cycle constructed using the three elevation angles. The corresponding best-fit plane is shown in gray. c) Side view of the cyclogram and best-fit plane to illustrate the normal distances between the trajectory and the best-fit plane used for computing the RMSE in linear dependency analysis.} 
  \label{fig:methods}
\end{figure*}

\subsubsection{Experimental Tasks}
The experimental tasks are designed to compare the natural biomechanical responses and the progression of these responses during the three scenarios; natural gait without the exoskeleton (No-Exo), exoskeleton-assisted locomotion (Exo-Assisted), and with-exoskeleton walking without any assistance nor resistance from the exoskeleton (Exo Zero-Torque). All tasks are performed on the treadmill at the participant-specific preferred walking speeds. To capture the early stages of natural motor adaptation to exoskeleton assistance in novice participants, the order of the trials was not randomized.

To establish the baseline kinematics and metabolic response data for nominal walking, each participant began the experiment by walking on the treadmill without the exoskeleton for 6 minutes at their predetermined walking speed. The 6-minute duration was selected on the basis of treadmill walking adaptation study \cite{6minTreadmill} and results from preliminary pilot trials. A minimum of 5 min break was provided between each experimental task to ensure reliable measurement of transient metabolic responses during adaption phases.

The participants were then donned with the exoskeleton to perform 25 minutes of Exo-Assisted treadmill walking at the same predetermined walking speed as before. The 25-minute duration was chosen to encompass the time required for adaptation in exoskeleton-assisted locomotion reported in the literature, defined according to the stabilization of metabolic and kinematic responses \cite{25min,25min_2}. The exoskeleton provided bidirectional torque assistance throughout the assisted walking trials, assisting in both plantar flexion and dorsiflexion. Torque assistance profile was prescribed using the default Proportional Joint Moment Controller (PJMC) \cite{spark}. The peak torque values were selected relative to the participant's body weight: 0.15-0.21 Nm/kg for plantar flexion assistance and 0.03-0.05 Nm/kg for dorsiflexion assistance – comparable with previous studies using this device \cite{Orehhov2020,fang2022}.

Lastly, participants performed 6 minutes of Exo Zero-Torque treadmill walking. In zero-torque mode, the exoskeleton regulates its applied torque at the ankle joint to zero. In this configuration, the zero-torque controller actively compensates for motor inertia, refraining from providing any assistance or resistance during the natural ankle plantar- and dorsiflexion motions. This allowed for a detailed examination of the physical implications caused by mechanical characteristics of the human-exoskeleton pair, the added weight, and the joint limits imposed by the exoskeleton.

\subsection{Post-Processing and Analysis}
\subsubsection{Kinematics-Based Gait Segmentation}
While markerless motion capture system provides reliable kinematics information, application of automatic gait segmentation methods using markerless motion capture data is not yet supported and remain underexplored. To extract key gait events from only the lower limb kinematics data, where each joint trajectory is concatenated into a continuous dataset per task, a reliable segmentation method was required. Although the reliability of gait event detection methods may vary depending on the specific experiment setup, two kinematics-based methods demonstrated highly reliable detection of heel strike and toe-off in this study. 

For toe-off detection, the maximum horizontal distance in the anterior-posterior (AP) axis from the toe landmark to the pelvist landmark provided reliable results \cite{toeOffDetection}. While the same method is also often used for heel strike detection using the hell landmark instead, the detected timestamps were always 75-100 ms (9-12 frames in 120 fps) too early. This occurred due to swing foot retraction: the heel swings downwards just moments before the heel strike, which reduces the horizontal distance between the heel and the pelvis. Instead, using the maximum contra-lateral hip extension resulted in error-free heel strike detection \cite{heelStrikeDetection}. With the reliable heel strike and toe-off detection methods, the gait kinematics data were segmented into two phases: the stance phase and the swing phase.

\begin{figure*}[!ht]
  \centering 
  \includegraphics[width=1.0\linewidth]{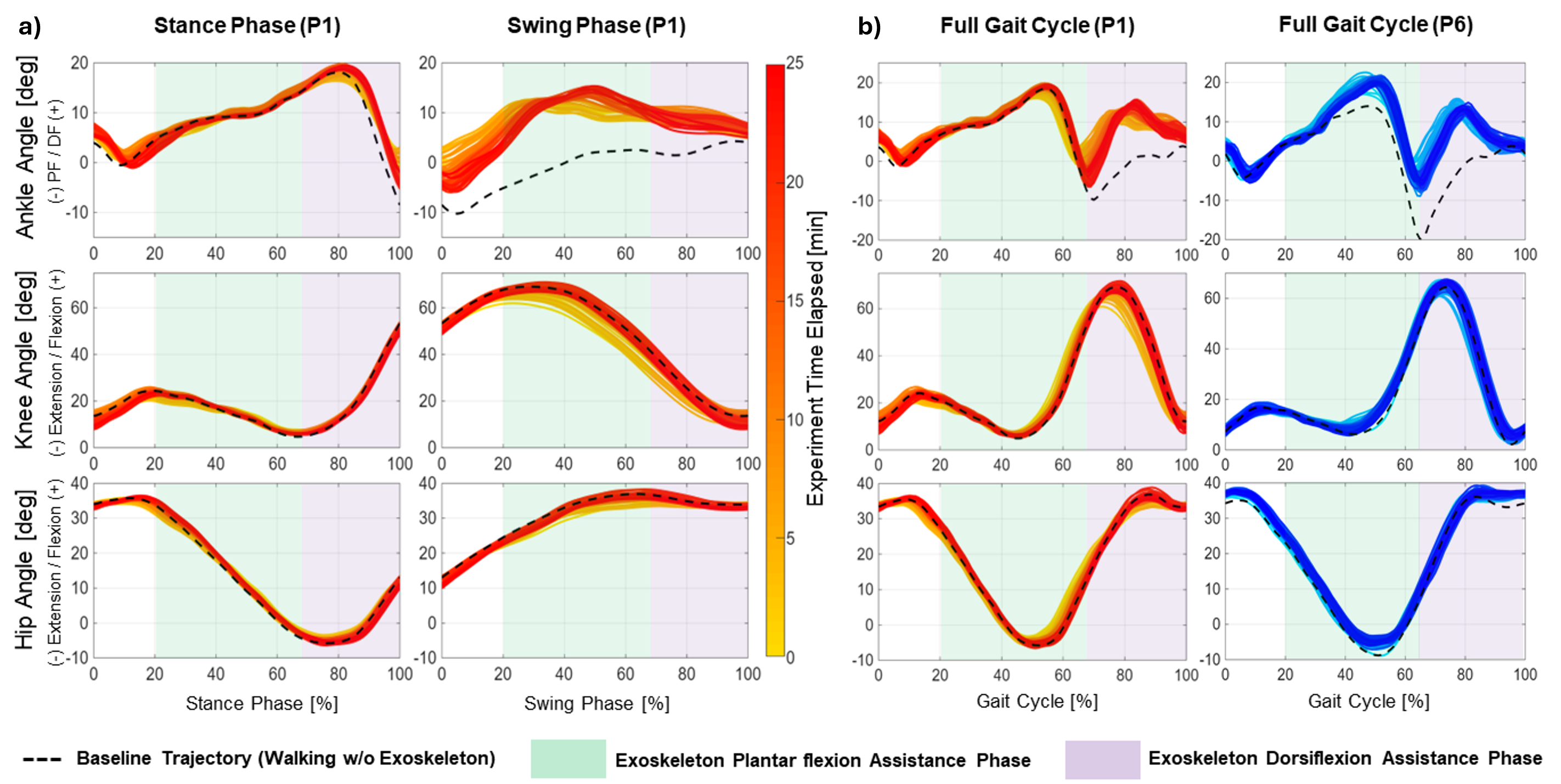}
  \caption{Temporal evolution of joint trajectories throughout the Exo-Assisted walking, illustrating the kinematic adaptation process. a) Comparison of the lower limb joint trajectories in stance and swing phases of P1. The color gradients indicate the temporal progression from the start (yellow) to the end (red) of the experiment. The dashed black line represents the baseline trajectory. Green and purple shaded regions indicate the timing of exoskeleton assistance, corresponding to plantar flexion and dorsiflexion assistance phases, respectively. b) Comparison of full gait cycle trajectories of P1 and P6. The blue trajectories correspond to joint trajectories of P6, showcasing distinct kinematic adaptation pattern in the ankle joint.}
  \label{fig:Traj}
\end{figure*}
\subsubsection{Lower Limb Kinematics}
Rather than focusing only on the statistical inferences, the adaptation strategies were studied individually per participant. Instead of analyzing the lower limb kinematic trajectories per gait cycle, the gait cycles were grouped into static 15-second time windows for averaging. The effects of different time window durations are elaborated further in the discussion. 

The course of the changes in lower limb kinematics were assessed by closely examining the evolution of the ankle, knee, and hip joint trajectories in the sagittal plane during the exoskeleton use. Specifically, the temporal evolution of joint trajectories was visualized to qualitatively assess the kinematic convergence features across different walking tasks and to study the kinematic implications associated with the donning and the usage of the exoskeleton. Then, using the averaged trajectory of the final three minutes of the experiment as the reference, the joint-level inter-trajectory RMSE was computed between the time-window averaged trajectories and the reference trajectories to quantitatively assess the temporal progress of kinematic adaptation. Specifically, a 2-degree kinematic convergence threshold was formulated using the RMSE observed in No-Exo task as a reference convergence criteria. In addition, the non-normalized trajectory data were utilized to analyze the changes in temporal parameters of gait, including the stance phase duty factor and stride duration. 

\subsubsection{Inter-Joint Coordination}
The inter-joint coordination was analyzed via computing the linear dependency among ankle, knee, and hip movements in the sagittal plane, and monitoring the changes in this linear dependency throughout the experiment. Specifically, the absolute elevation angles were defined for the lower limb segments: foot, shank, and thigh, to construct a 3D cyclogram. The coefficients of the best-fitting 2D planes were then derived for each discretized time window, using the total least-squares approach via singular value decomposition (SVD). Similarly, the RMSE of normal distance to the best-fit plane was also computed for every discretized time window, as demonstrated in Fig. \ref{fig:methods}c. Based on the formulation of the SVD, the RMSE of normal distance to the best-fit plane was simply derived from utilizing the smallest singular value: 

\begin{equation}
    LD\ RMSE\ =\sqrt{\frac{1}{N}\sum_{i=1}^{N}d_i^2}=\frac{\sigma_3}{\sqrt N}    
\end{equation}

Where $N$ is the number of data points in the cyclogram, $d_i$ is the perpendicular distance of an individual point to the fitted plane, and $\sigma_3$ is the smallest singular value from the singular value decomposition. This linear dependency RMSE metric was used for observing how the inter-joint coordination evolved throughout the exoskeleton use and to compare how the responses varied across different walking tasks.

\begin{figure*}[!t]
  \centering 
  \includegraphics[width=1.0\linewidth,trim=2 2 2 2,clip]{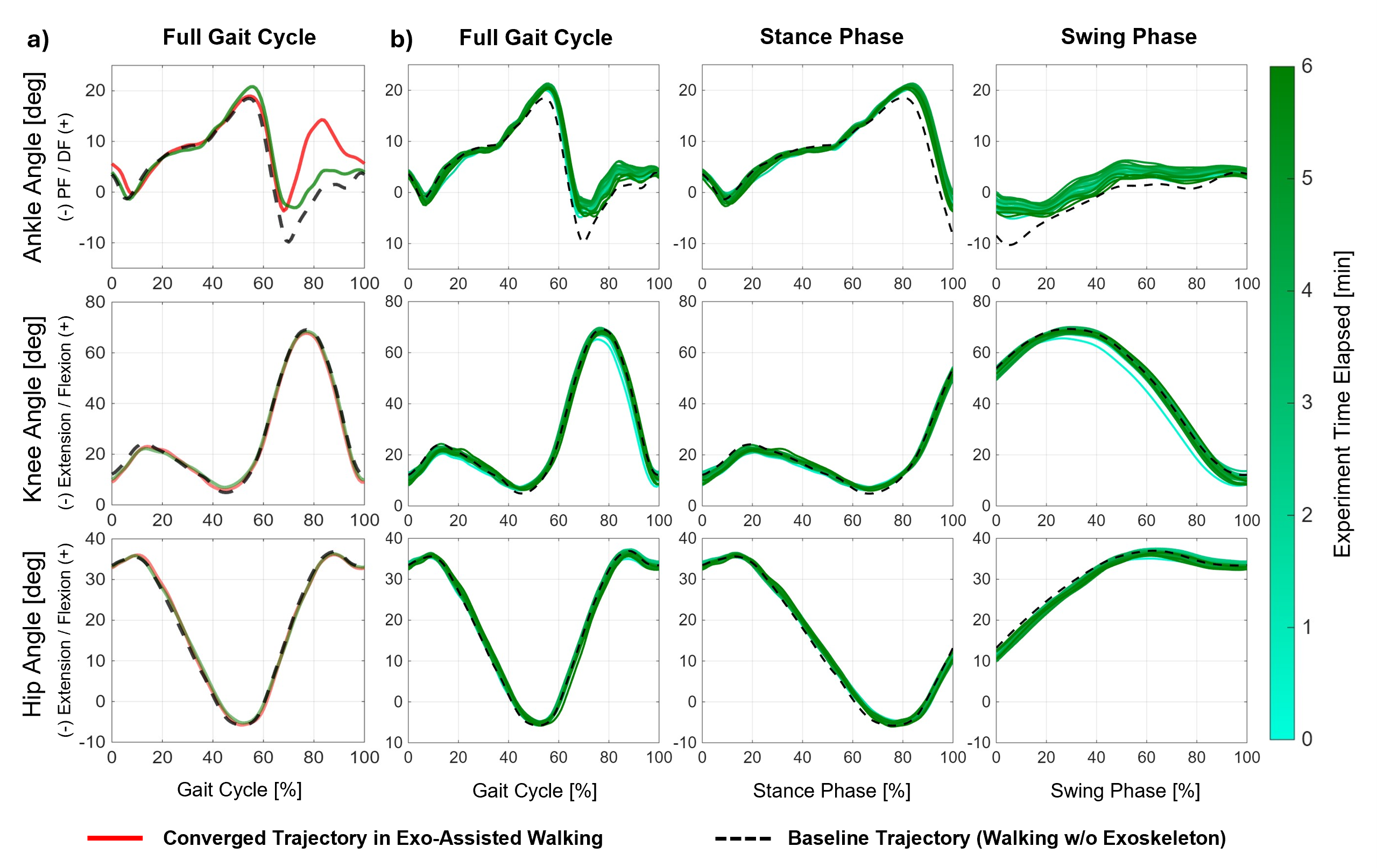}
  \caption{Lower limb trajectories compared to baseline trajectory of P1 data. a) Direct comparison of the converged Exo-Assisted walking (in red) and the converged Exo Zero-Torque walking (in green) with the baseline trajectory (in dashed black). b) Temporal evolution of joint trajectories during the Exo Zero-Torque walking. The color gradients indicate temporal progression from the start (teal) to the end (dark green).}
  \label{fig:TrajZT}
\end{figure*}

\subsubsection{Relative Change in Metabolic Cost of Transport}
The portable metabolic measurement system was calibrated and prepared prior to the experiment. The collected respiratory response data were post-processed using the Brockway equation to derive the metabolic response throughout the experiment \cite{brockwayEquation}. Given that every participant performed walking trials at a fixed walking speed, the metabolic cost of transport (MCoT) was determined using the baseline metabolic rate collected during pre-experimental task, post-processed metabolic rate, the subject-specific walking speeds, and the task-specific total weights. Specifically, the added weight of the exoskeleton was taken into the account when computing the MCoT during Exo-Assisted and Zero-Torque trials. Leveraging the stabilized metabolic response in the last minute of the No-Exo walking, the subject-specific relative change in MCoT was computed by finding the difference in No-Exo and Exo-Assisted and normalizing to the No-Exo response.

\subsubsection{Subjective Perception of Adaptation}
To observe the subjective perception of adaptation to exoskeleton use, the study also included a simple subjective task during the exoskeleton-assisted walking task. Specifically, the participants were asked to raise their arm at any point during the 25-minute exoskeleton-assisted treadmill walking task if they perceived they had adapted to the exoskeleton. To avoid introducing any bias, participants were reminded of their option to refrain from raising their arm should they not feel adapted to the exoskeleton. Furthermore, no explicit definition of adaptation was imposed to the participants.

\begin{figure*}[t]
  \centering 
  \includegraphics[width=0.99\linewidth]{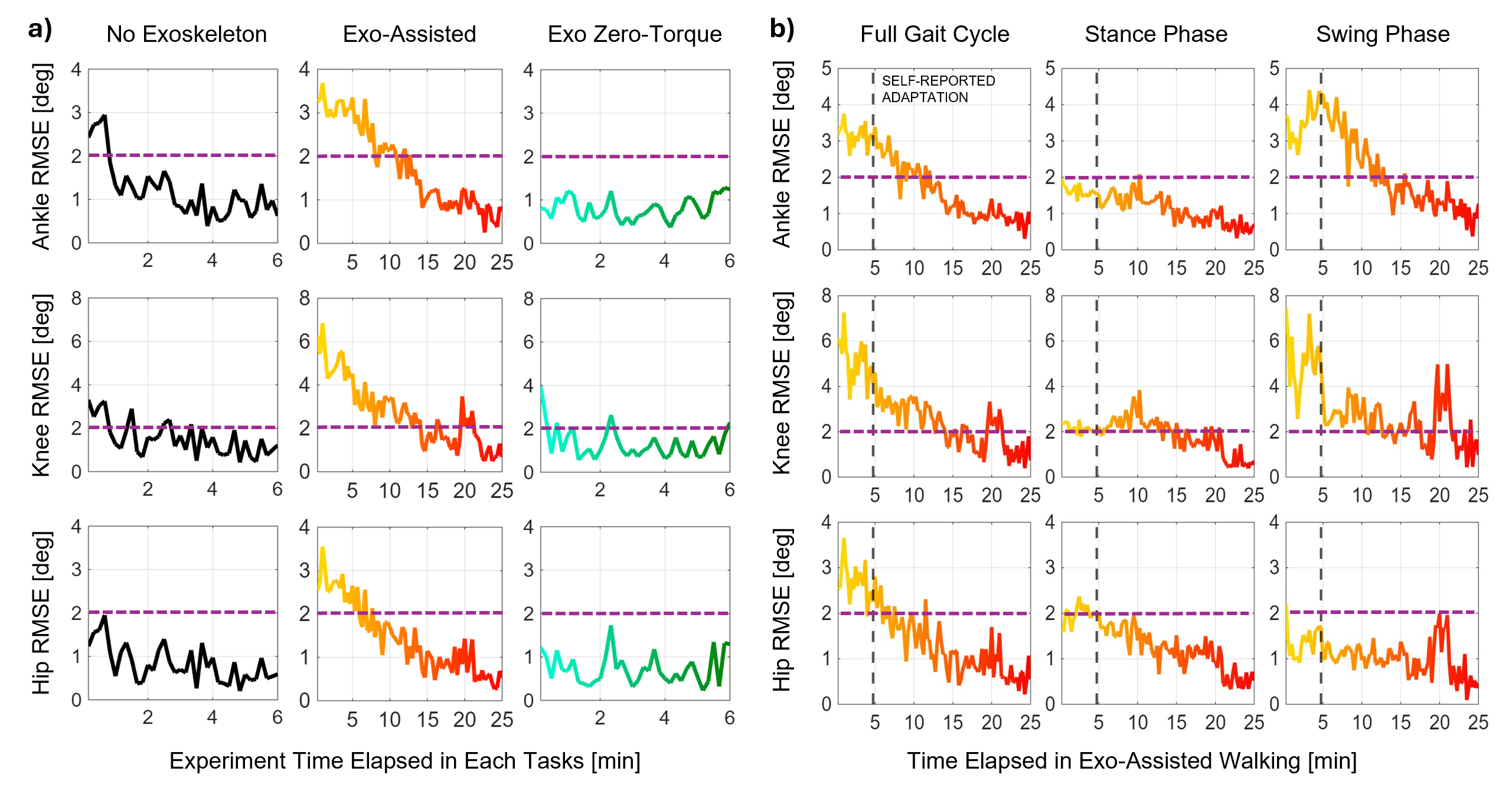}
  \caption{Joint-level inter-trajectory RMSE for P1. a) RMSE responses across the three experimental tasks. b) Per-phase RMSE in the Exo-Assisted task, highlighting the higher inter-trajectory variances during the swing phase. The black vertical dashed lines represent the self-reported adaptation timings, and the purple horizontal dashed lines show the 2-deg convergence threshold.}
  \label{fig:RMSE}
\end{figure*}

\section{Results}
With the individual responses as the focus of the study, detailed analysis of a sample participant (P1) is provided in this section. The figures are populated individually for all participants and are included in the Supplementary document S1. A detailed comprehensive inter-participant analysis and comparisons are explored further in the discussion section.

Note that P3 and P8 were unable to complete the Exo Zero-Torque task due to technical limitations of the exoskeleton, and the indirect calorimetry for P7 and P9 resulted in inaccurate measurements due to premature sensor failure. Consequently, the kinematic and spatio-temporal gait analyses were conducted across N=9 participants for No-Exo and Exo-Assisted walking, and N=7 participants for the Exo Zero-Torque task. Similarly, the MCoT responses were computed across N=7 participants.

\subsection{Kinematics Analyses}
The temporal evolution of the joint trajectories during the Exo-Assisted walking is presented in Fig. \ref{fig:Traj}, illustrating how the joint trajectories changed throughout the course of the adaptation process during exoskeleton-assisted treadmill walking. For all participants, the largest kinematic deviations from the reference trajectory of No-Exo walking occurred at the ankle joint, where active exoskeleton assistance was applied. Although all participants eventually demonstrated kinematic convergence according to the observed evolution of joint trajectories, the deviation characteristics varied across individuals. 

For instance, P1 maintained stance phase peak dorsiflexion angles close to the baseline throughout the experiment, whereas P6 exhibited persistently increased dorsiflexion. Nevertheless, several common patterns were observed across participants. For example, both P1 and P6 initially showed substantially reduced peak plantar flexion angle at the start of the swing phase relative to the baseline trajectory. Furthermore, the peak dorsiflexion angle during swing remained consistently higher than that of the reference trajectory for all participants, although the timing of convergence differed substantially (see Supplementary Document S1). 

Analyzing the temporal evolution of joint trajectories individually per participant revealed the inter-joint differences in adaptation characteristics. For P1, for example, the hip trajectories during the Exo-Assisted task were promptly well-aligned with the reference trajectory of No-Exo walking task from the earlier phase of the experiment, where the observed deviations in the hip ROM were minor relative to the large initial deviations shown in the ankle trajectories. In addition, the hip trajectories swiftly reached steady-state where the inter-trajectory deviation became minimal, suggesting that a rapid kinematic convergence was acquired in the hip joint. In contrast, the ankle and the knee trajectories exhibited larger inter-trajectory deviation overall, in which the relative differences were more apparent and showcased slower convergence during the swing phase. 

In contrast to Exo-Assisted walking, the temporal evolution of joint trajectories during the Exo Zero-Torque task (Fig. \ref{fig:TrajZT}) showed minimal deviations from the baseline trajectories throughout the experiment. Overall, the knee and hip trajectories remained closely aligned with the baseline trajectory, resulting in minimal inter-trajectory deviations throughout the experiment. While the ankle trajectories exhibited larger deviations from the baseline trajectory compared to the other joints, the overall inter-trajectory deviations were smaller than that of Exo-Assisted walking. The most notable deviation from the baseline trajectory was the increase in peak dorsiflexion angle during the stance phase and the decrease in peak plantar flexion angle at the onset of the swing phase. The increase in peak dorsiflexion during stance phase was observed in all 7 participants that completed Exo Zero-Torque task. However, the magnitude, inter-trajectory deviation, and timing of the observed kinematic convergence varied substantially across participants. 

Further insight regarding joint-level kinematic convergence behavior was obtained across different walking tasks by analyzing the inter-trajectory RMSE, which computes the degree of inter-trajectory deviation between trajectories over time. As shown in Fig. \ref{fig:RMSE}a, rapid convergence was observed in both No-Exo and Exo Zero-Torque tasks where the RMSE of all joint trajectories stabilized below the 2-degree convergence threshold within the first few minutes. Specifically, all 9 participants achieved this convergence threshold across all joints in No-Exo task, whereas all but 2 participants (P2 and P4) achieved the same converging response during the Exo Zero-Torque task. For these 2 participants, only the knee inter-trajectory RMSE remained above the threshold, while the hip and the ankle RMSE showcased rapid convergence below the threshold. 

\begin{table}[t]
    \centering
    \caption{Summary of convergence timings across participants. Timestamps denote kinematic convergence, bracketed values indicate the duration of deviation from the convergence criterion (in minutes).}
    \label{table:table2}
    \renewcommand{\arraystretch}{1}
    \begin{tabular}{|c|c|c|c|c|}
        \hline
        Subject & Joint & No-Exo & Exo-Assisted & Exo Zero-Torque \\
        \hline
            & Ankle & 0.87 & 12.08 & 0.25 \\
        P1  & Knee  & 0.86 & 14.02 (21.37-end) & 0.42 \\
            & Hip   & 0.25 & 11.59 & 0.25 \\
        \hline
            & Ankle & 0.25 & 12.91 (14.55-18.45) & 0.25 \\
        P2  & Knee  & 1.64 & 13.65 (14.35-19.36) & 0.25 (1.19-5.09) \\
            & Hip   & 0.25 & 3.12 & 0.25 \\
        \hline
            & Ankle & 0.25 & 18.63 & N/A \\
        P3  & Knee  & 1.11 & 14.81 (19.19-20.69) & N/A \\
            & Hip   & 0.25 & 3.95 & N/A \\
        \hline
            & Ankle & 0.25 & 0.88 & 0.25 \\
        P4  & Knee  & 0.25 & 0.80 (3.55-21.69) & 0.25 (5.61-end) \\
            & Hip   & 0.25 & 0.45 (4.10-4.48) & 0.25 \\
        \hline
            & Ankle & 0.25 & 8.77 & 0.30 \\
        P5  & Knee  & 0.85 & 14.92 (17.64-18.10) & 1.48 \\
            & Hip   & 0.25 & 7.65 & 0.37 \\
       \hline
            & Ankle & 0.25 & 5.36 & 0.25 \\
        P6  & Knee  & 0.25 & 5.80 (8.41-18.76) & 1.15 \\
            & Hip   & 0.25 & 0.91 & 0.32 \\
        \hline
            & Ankle & 0.25 & 4.01 (13.93-14.83) & 0.30 \\
        P7  & Knee  & 0.25 & 3.71 (8.32-21.97) & 2.87 \\
            & Hip   & 0.25 & 1.76 & 0.25 \\
        \hline
            & Ankle & 0.25 & 18.78 & N/A \\
        P8  & Knee  & 0.25 & (Start-end) &  N/A \\
            & Hip   & 0.27 & 12.56 & N/A \\
        \hline
            & Ankle & 0.25 & 7.88 & 0.68 \\
        P9  & Knee  & 0.50 & 12.42 (13.49-end) & 1.13 \\
            & Hip   & 0.38 & 4.94 & 0.25 \\
        \hline
    \end{tabular}
\end{table}

Utilizing the same 2-degree inter-trajectory RMSE convergence criterion, Exo-Assisted walking required longer adaptation periods overall. For P1, for example, the ankle and hip inter-trajectory RMSE stabilized below the threshold within the within the first 10 minutes, while the RMSE convergence occurred around the 15th minute mark for the knee trajectories. Even so, the knee trajectories occasionally exhibited deviations above the convergence threshold in the 19$^{th}$ minute mark. 

A summary of kinematic convergence timings across participants is as provided in Table \ref{table:table2}. Despite the inter-participant differences in deviation characteristics, all participants did acquire convergence for the ankle and the hip joints during Exo-Assisted walking, albeit the timing and the rate of convergence were vastly different across participants. Interestingly, all participants except P3 failed to achieve inter-trajectory RMSE convergence in the knee trajectories.

The phase-specific RMSE of the Exo-Assisted walking task is as shown in Fig. \ref{fig:RMSE}b. The results clearly indicate greater trajectory variance during swing, particularly for the ankle and the knee joints. This pattern of ankle and knee trajectories showcasing higher inter-trajectory RMSE in swing phase was consistent across all participants. 
%
\begin{figure}[!t]
  \centering 
  \includegraphics[width=1.0\linewidth,trim=1 1 1 1,clip]{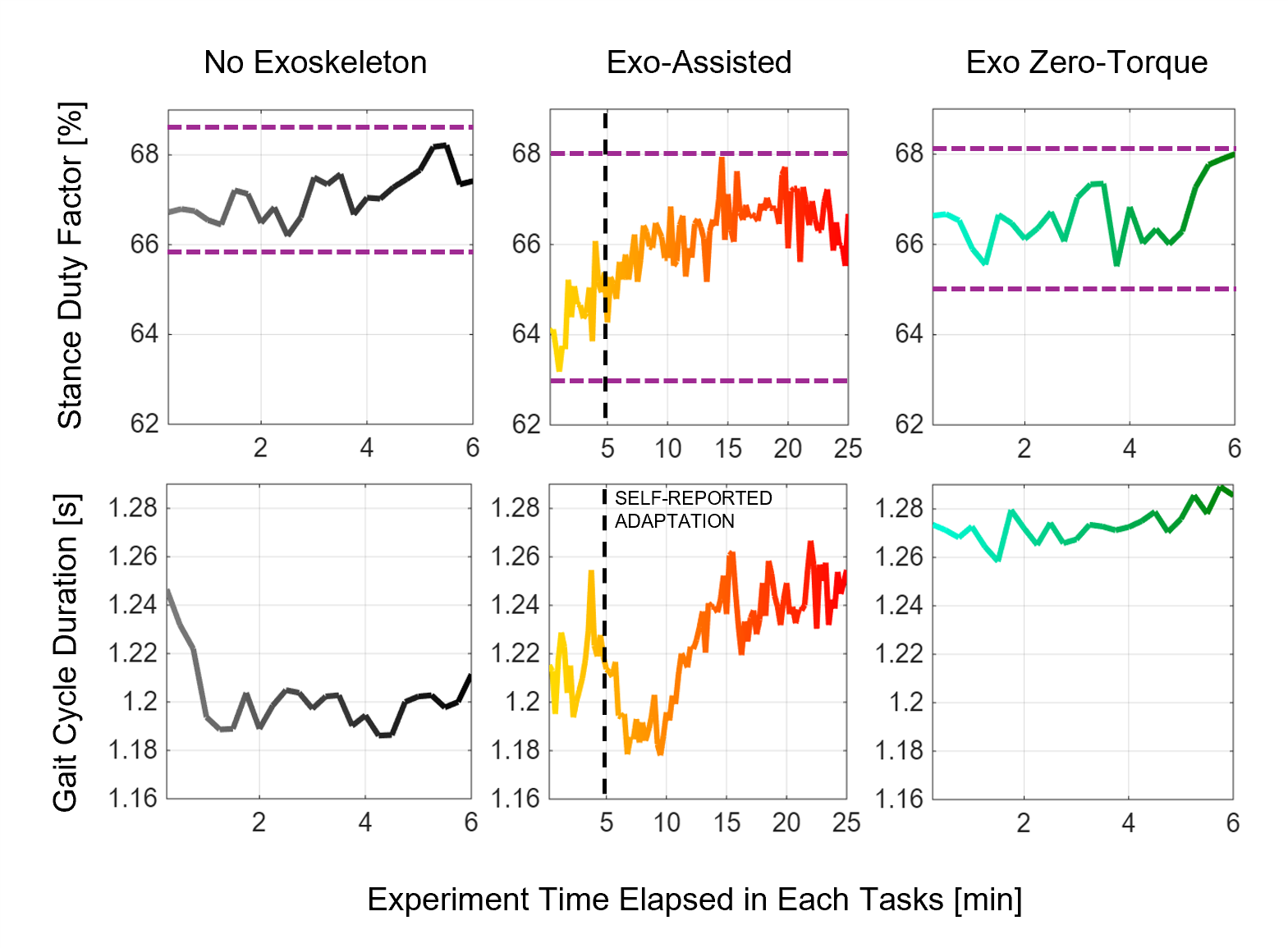}
  \caption{Temporal evolution of stance phase duty factor (top) and the gait cycle duration (bottom) for P1 across the three walking tasks. Dashed purple lines denote the $\pm$3\% (in No-Exo and Exo Zero-Torque) and $\pm$6\% (Exo-Assisted) duty factor envelopes encapsulating all participant responses.}
  \label{fig:TG}
\end{figure}

\subsection{Temporal Gait Characteristics}
Fig. \ref{fig:TG} depicts how the temporal gait characteristics changed throughout the experiment. The changes in stance phase duty factors remained minimal in both the No-Exo and the Exo Zero-Torque conditions, in which the duty factor varied within a 3\% envelope throughout the two experimental tasks for all participants. For the Exo-Assisted walking, this envelope was increased to 6\% but the magnitude of stance phase duty factor were always on par with No-Exo walking tasks. 

In addition to the changes in phase duration, changes in stride duration also remained minimal during both the No-Exo and Exo Zero-Torque tasks. In the Exo-Assisted task, the stride duration acquired temporary convergence after undergoing fluctuating response in the first 7 minutes, which then it inclined by 4.6\% from the temporarily-converged level. Interestingly, the mean stride duration during the Exo Zero-Torque walking was increased by 5.8\% relative to that of No-Exo walking. Overall, the changes in stride duration were minimal in No-Exo and Exo Zero-Torque tasks throughout the experiments where the maximum observed range of stride duration was 0.07 seconds per gait across all participants, while the changes were more drastic in the Exo-Assisted walking with a maximum range of 0.15 seconds per gait.


\subsection{Inter-joint coordination over time}
\begin{figure*}[!ht]
  \centering 
  \includegraphics[width=1.0\linewidth]{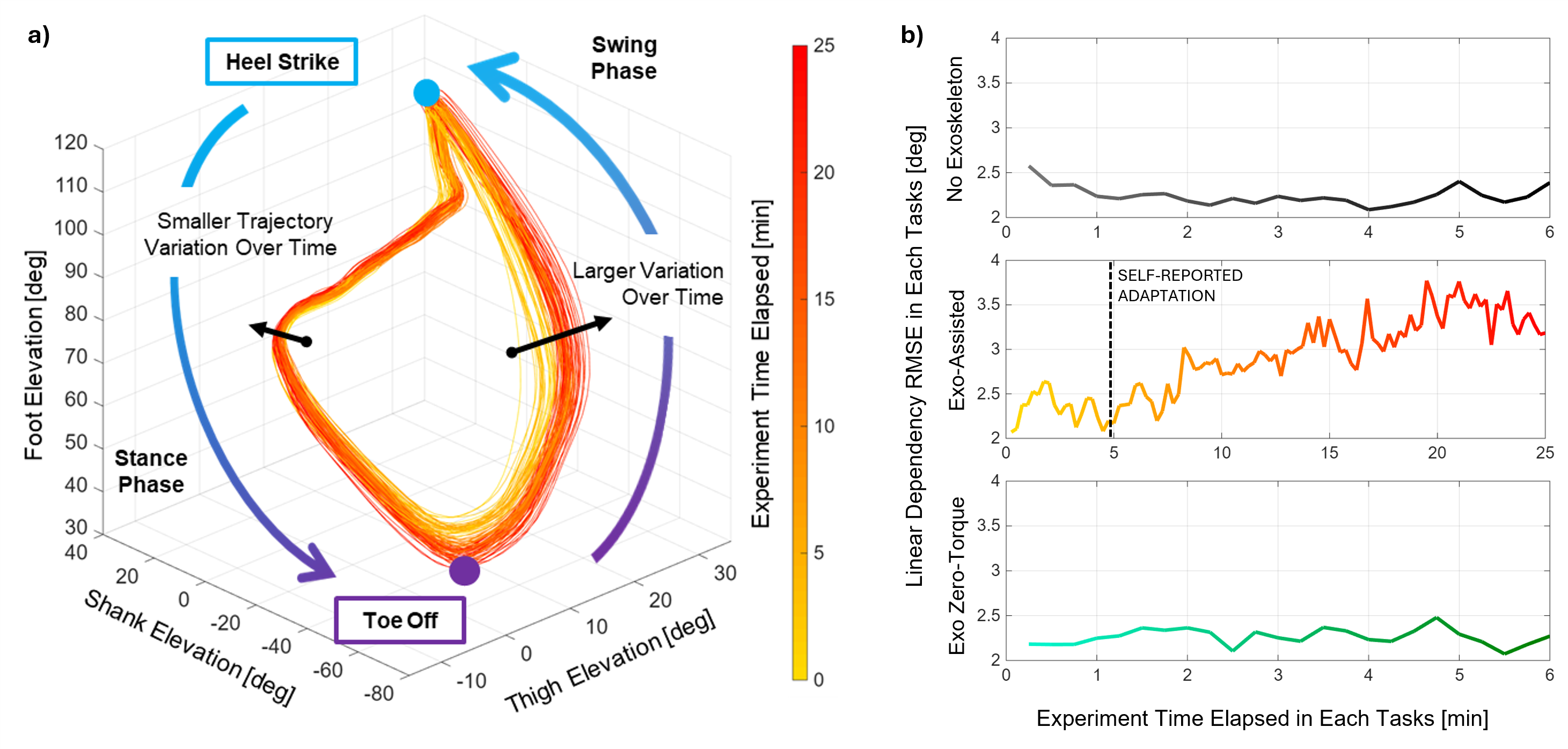}
  \caption{Temporal evolution of Inter-joint coordination and linear dependency over time. a) Overlaid 3D cyclograms of the lower limb elevation angles during the Exo-Assisted walking for P1, illustraing the progressive changes in inter-joint coordination over the course of the experiment. Reduced trajectory dispersion is observed during the stance phase, whereas larger inter-trajectory variability occurs during swing pahse. The overlapping trajectories in the later part of th experiment indicate convergence. b) linear dependency analysis using the RMSE of inter-joint cyclograms to their best-fitted planes in the three tasks.}
  \label{fig:LD_Combined}
\end{figure*}

Fig. \ref{fig:LD_Combined}a depicts the overlaid 3D cyclogram of the lower limb elevation angles during the Exo-Assisted walking, showcasing the progressive shift in the inter-joint angle relations. Similar to temporal evolution of joint trajectories (Fig. \ref{fig:Traj}), inter-joint coordination convergence was visually demonstrated by cyclogram progressively overlapping closer later in the experiment. Evidently, inter-trajectory variations in inter-joint coordination was larger during the swing phase of the gait cycle compared to the stance phase.

The changes in linear dependency among the lower limb segments were evaluated across the different experimental tasks and are as shown in Fig \ref{fig:LD_Combined}b. While the RMSE of the inter-joint trajectories to their respective best-fitted planes remained low during both the No-Exo and the Exo Zero-Torque walking tasks, the RMSE in Exo-Assisted walking showcased larger deviations from linear dependence. For P1 specifically, the linear dependency RMSE in Exo-Assisted walking was initially similar in magnitude to that of the No-Exo and Exo Zero-Torque cases, which then increased to higher RMSE value later in the experiment.

Although individual responses varied across participants, Exo-Assisted walking consistently demonstrated higher linear dependency RMSE values and larger deviations throughout the task relative to the No-Exo condition. Specifically, 8 participants had noticeably increased linear dependency RMSE values during the Exo-Assisted walking task, which reflects decreased linear dependency. As for the remaining participant, the RMSE in Exo-Assisted walking was comparable to that of the No-Exo walking task.

Concerning the differences in No-Exo and Exo Zero-Torque RMSE, only 1 of the 7 participants data (P1) showcased increased linear dependency RMSE in the Exo Zero-Torque task relative to No-Exo walking. While 3 participants' RMSE in Exo Zero-Torque were comparable to No-Exo condition (P1, P2, and P7), the other 3 participants' RMSE resulted in noticeably lower RMSE in Exo Zero-Torque task (P4, P5, and P6). 

\subsection{Group-Averaged Kinematics}
\begin{figure*}[t!]
  \centering 
  \includegraphics[width=1.0\linewidth]{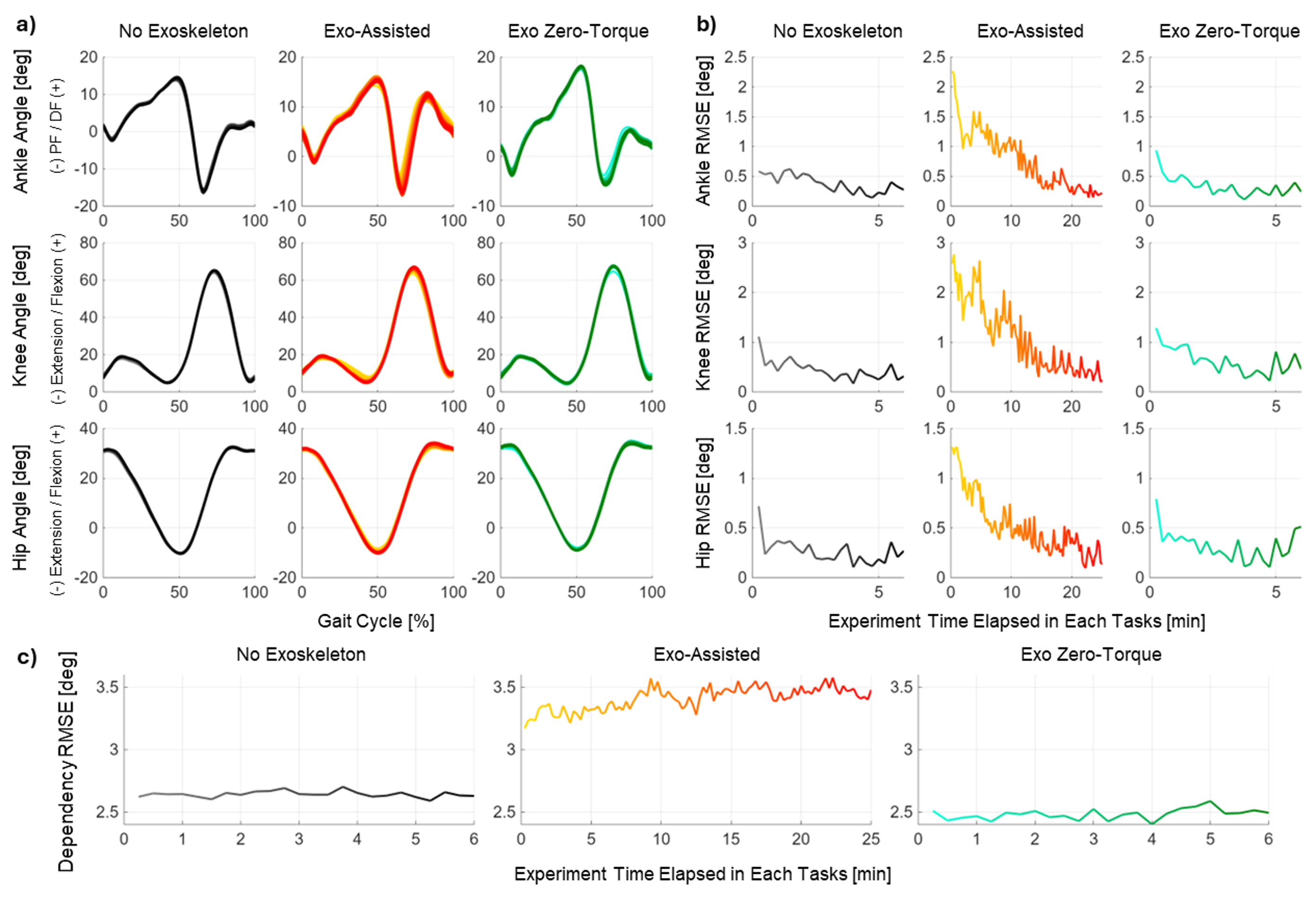}
  \caption{Group-level averaged kinematic responses. a) The lower limb joint trajectories. b) The joint-level inter-trajectory RMSE voer time, showcasing reduced overall RMSE due to variance-attenuating nature of averaging and representing a homogeneous adaptation pattern. c) The linear dependency RMSE.}
  \label{fig:stats} 
\end{figure*}

Fig. \ref{fig:stats} presents the same kinematic analyses obtained using the group-averaged trajectory data. Unlike the participant-level analysis, the group-averaged results exhibited minimal inter-trajectory RMSE across all walking tasks. Specifically, all inter-trajectory RMSE except for the knee joint during Exo-Assisted instantly fulfilled the 2-degree RMSE convergence at the start of the experiment. For the knee joint during Exo-Assisted, kinematic convergence was reported after the 8$^{th}$ minute.


\subsection{Metabolic Cost of Transport}
\begin{figure}[t]
  \centering 
  \includegraphics[width=1.0\linewidth]{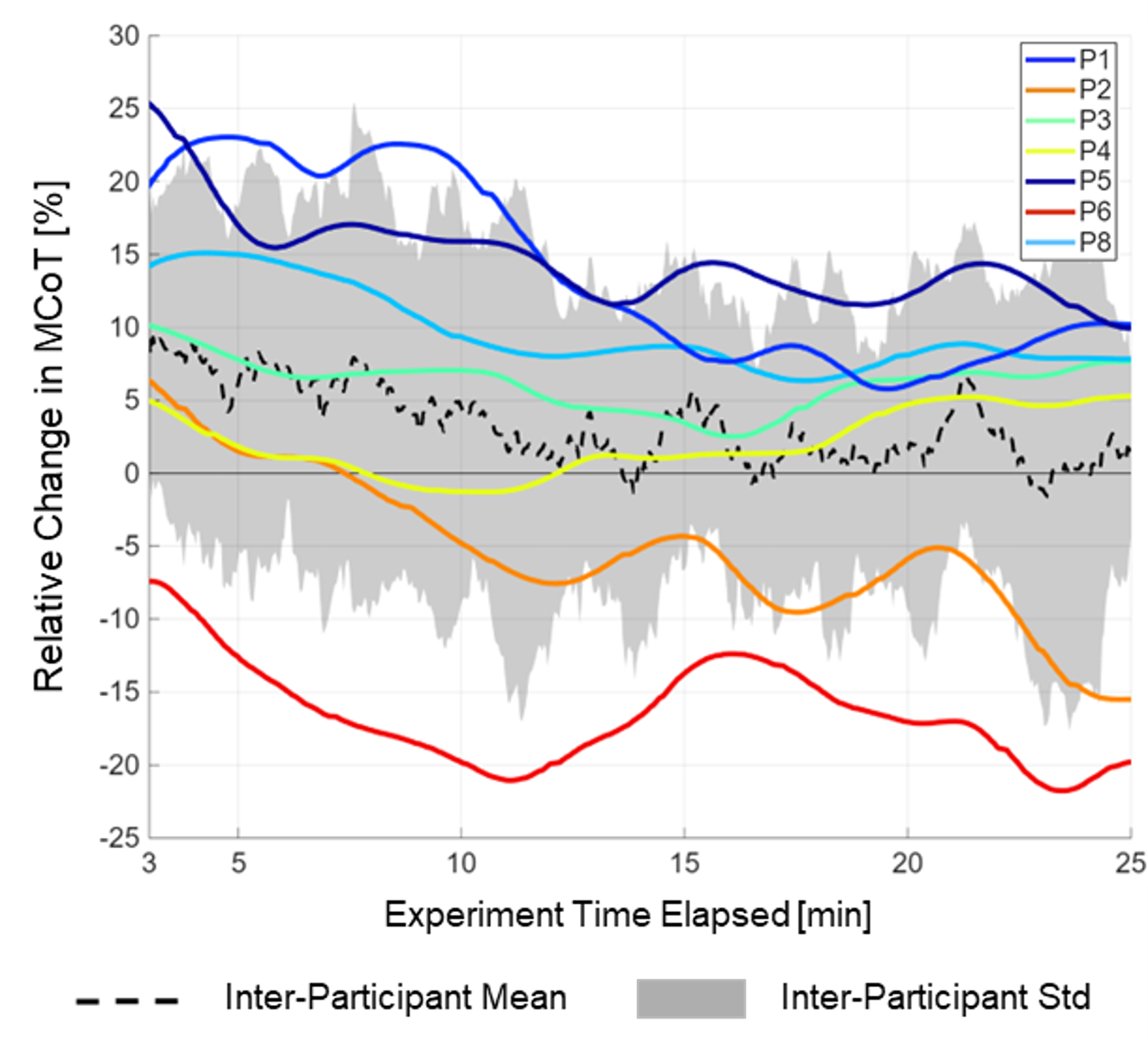}
  \caption{Relative changes in MCoT responses during Exo-Assisted task with respect to the No-Exo reference. Colored lines represent individual participant responses. The dashed black line and the shaded gray region represent the inter-participant mean and standard deviation, respectively.}
  \label{fig:MCoT}
\end{figure}

Fig. \ref{fig:MCoT} shows the participant-specific and group-average changes in the MCoT response during Exo-Assisted walking relative to the No-Exo task. On average, participants did not exhibit a reduction in MCoT, but rather an overall increase of 5\%. Although 5 out of 7 participants with metabolic response data showed a decreasing MCoT trend over time, responses were highly heterogeneous across participants, as reflected by the large standard deviation ranging from  6.89\% shown in the 18$^{th}$ minute to 18.17\% which occurred in the 11$^{th}$ minute, without an uniform trend observed. 

On the contrary, MCoT response during No-Exo task and Exo Zero-Torque rapidly reached to a steady-state within the first few minutes of the walking task.

\subsection{Subjective Perception}
Only 5 out of 9 participants reported experiencing adaptation to the exoskeleton-assisted walking (P1, P2, P5, P7, and P9). The self-reported adaptation timings are incorporated throughout the presented figures as vertical dashed black lines. Among those who did, the perceived timing of adaptation varied substantially, with a mean of 545.60 $\pm$ 448.65 seconds into exoskeleton walking task. Although these participants showcased rapid decline in swing phase joint-level inter-trajectory RMSE when they reported being adapted to exoskeleton use, this self-reported adaptation did not necessarily correlate with the global minimum of RMSE, nor did it align with where the fastest decline in the RMSE response occurred. Overall, participants' subjective reports did not consistently align with the objective indicators of adaptation, including kinematics and metabolic cost responses. For instance, participant P1 reported feeling adapted to exoskeleton assistance at 4 minutes and 45 seconds into the walking task, but the timing did not align with any of the observed kinematic or metabolic adaptation behavior.

\section{Discussion}
Based on the presented results, motor adaptation to exoskeleton-assisted locomotion appears to be a gradual progress, with substantial inter-participant differences. Although most participants eventually attained kinematic convergence according to the adopted metrics, both the timing of the convergence and the magnitude of kinematic deviations varied considerably across individuals. This variability highlights the inherently individualized nature of adaptation, emphasizing the importance of individual analysis in complex pHEI such as exoskeleton-assisted locomotion. 

In addition, kinematic adaptation occurred asynchronously across the lower limb joints for all participants. Specifically, the kinematic convergence of the hip, knee, and ankle trajectories did not occur simultaneously, even among participants who acquired rapid kinematic adaptation. This indicates that the underlying factors governing the rate of kinematic adaptation are not homogeneous across lower limbs, where the attainment of kinematic convergence in one joint does not imply simultaneous convergence in other joints. These inter-joint differences were amplified during the Exo-Assisted walking compared to No-Exo and Exo Zero-Torque tasks, highlighting that the addition of torque assistance drastically alters the lower limb synchrony and adaptation dynamics.

In Exo-Assisted walking with an active ankle exoskeleton, the ankle joint is the primary location for the pHEI since it is where the assistive torque is directly applied. Consequently, ankle kinematics are expected to be particularly sensitive to external loads, and hence exhibit slow kinematic convergence. However, the knee and the hip joints also exhibited much slower kinematic convergence during the Exo-Assisted task relative to No-Exo and Exo Zero-Torque walking, with the knee joint often deviating above the threshold during Exo-Assisted walking. As participants interact with the exoskeleton, the neuromuscular system needs to adjust its control strategy to incorporate the provided assistance. In this context, the slow kinematic convergence observed in the knee and hip reflects both the sensitivity of joints indirectly affected by the exoskeleton and the persistent adjustments across lower limbs during adaptation.

The larger magnitude and higher fluctuations of the inter-trajectory RMSE during the swing phase are especially intriguing, considering that the torque applied by the exoskeleton for dorsiflexion assistance during swing is considerably smaller than the plantar flexion assistance during stance. During stance, ground contact forms a closed kinetic chain that constrains ankle motion through ground reaction forces. In contrast, there exists no ground contact for the foot in swing phase, resulting in the human-exoskeleton pair effectively operating as an open kinetic chain. Consequently, ankle kinematics are inherently more sensitive to exoskeleton assistance-induced perturbations and the dynamic pHEI, contributing to larger trajectory fluctuations observed during swing.

There were no distinguished changes observed in the temporal aspect of gait, aside from the increased signal variance envelopes in Exo-Assisted walking. This is likely due to treadmill walking at a fixed speed, which inherently constrains the spatiotemporal variability compared to self-paced overground locomotion without exoskeletons \cite{lee2024}. Even though the treadmill speed was set according to individual walking preferences, having to maintain a constant walking speed to keep up with the treadmill constrained participants' ability to freely modify the temporal gait parameters, limiting the potential deviations during the adaptation process. It is likely that experiments conducted under a self-paced or overground walking condition would reveal more pronounced adaptation features in the context of temporal gait features, as participants would have more freedom to adjust their cadence and other gait characteristics.

Although the overall trajectory deviation from the baseline in Exo Zero-Torque walking was minimal relative to the more pronounced deviations observed during the Exo-Assisted walking, notable changes in ankle kinematics were observed despite the absence of exoskeleton assistance. Given that the exoskeleton actively refrained from providing assistance or resistance against the participant's movement during the Exo Zero-Torque walking, these changes are primarily attributed to the biomechanical characteristics of the human-exoskeleton interface. Specifically, the only factors introduced in the Exo Zero-Torque walking compared to No-Exo waking are the human-exoskeleton physical interaction attributes including non-anthropomorphic footplate, fixated ankle rotation axis, and constraints on ankle inversion and eversion. Although passive, these elements may influence the natural gait patterns.

However, it should also be noted that in Exo Zero-Torque mode, a non-zero lag is always present in the control response, as the device must detect torque changes via the sensor before compensating, resulting in unavoidable residual interaction torques. Hence, the dynamic characteristics of the control loop should not be neglected in the interpretation of the observed Exo Zero-Torque walking data.

While the inter-trajectory RMSE revealed valuable insights regarding the individual adaptation behavior, the analysis is limited to the joint-level information and does not fully encompass the details regarding how lower limb joints coordinate with one another throughout the adaptation process. The linear dependency analysis complemented such downside of joint-level metrics by encapsulating the characteristics of how inter-joint coordination patterns evolved throughout the pHEI adaptation process. 

In the proposed formulation, an increase in lower limb linear dependency RMSE corresponds to the increased deviation of the angle-angle trajectories from their best-fitted plane. In the context of exoskeleton-assisted locomotion, this does not necessarily indicate whether inter-joint coordination improves or degrades over time. Rather, it reflects an increasing linear independence among the lower limb joints, driven by the external control inputs introduced to the exoskeleton user that causes lower limb joints to deviate from the natural coordination patterns driven by the central nervous system (CNS) \cite{hicheur2006}. This interpretation aligns with exisint literature demonstrating that the addition of active joint assistance alters natural inter-joint coordination patterns \cite{lora-millan2022}.

In Exo-Assisted walking, the assistive torque profile provided by the exoskeleton is the only independent variable added to the human-exoskeleton pair. This is then the external variable in which the participants have to adapt to over the course of the experiment. In this regard, the changing yet converging linear dependency patterns can be interpreted as participants progressively adapting to admit the exoskeleton as part of their locomotion system. This interpretation also explains why there had not been any major changes in inter-joint linear dependency during No-Exo and Exo Zero-Torque walking trials, as there had not been torque assistance forcefully driving the inter-joint coordination to deviate from the natural patterns. In fact, the linear dependency RMSE in Exo Zero-Torque walking was even lower than than that of No-Exo walking for three participants, which may be caused by the kinematic constraint of the exoskeleton construction reducing the joint motions in non-sagittal plane such as ankle inversion and eversion. Given the participant-dependent nature of these responses, capturing human-exoskeleton interface dynamics in larger cohorts would be critical to unravel this phenomenon further.

Although group-level analyses can reveal the general trend observed from the per-participant analyses, the individualized nature of adaptation may be overlooked when relying solely on group-averaged results. Specifically, group-averaged kinematic trajectories fail to capture inter-participant differences, such as variabilities in kinematic convergence timings of individual participant. Furthermore, the joint-level inter-trajectory RMSE also becomes substantially attenuated. This is because averaging kinematics data across participants inherently reduces variance by attenuating the uncorrelated deviations in the data, which results in kinematic responses that appear more stable or faster converging compared to the underlying individual responses. Therefore, this reduction in inter-trajectory RMSE in the group-averaged kinematics data suggests that the adaptation dynamics of exoskeleton use can be incorrectly generalized when relying only on the group-level analysis. 

Moreover, such variance reduction can create a statistical illusion that may lead to incorrect interpretations of adaptation dynamics, stability, or the perceived benefits from exoskeleton use. For example, participants who did not adapt to the exoskeleton, or who adapted differently, may be masked by the group average. This can potentially lead to misclassification and overestimation of the apparent treatment effect. This was clearly demonstrated in the relative change in MCoT, where the participants showcased wide range of non-converging responses throughout the experiment, whereas the mean response portrayed a singular, converged response. While all participants of this study were healthy young adults who demonstrated steady kinematic adaptation responses evident from the inter-trajectory RMSE and inter-joint coordination metrics, motor adaptation responses may vary significantly depending on the study population. Given the broad spectrum of neuromuscular impairments, patient populations may adopt diverse adaptation strategies characterized by varying levels of conservative exploration and exoskeleton reliance. Therefore, individualized assessment becomes even more critical when translating these findings to clinical settings.

Aside from individualized assessment of adaptation, analyzing the temporal progression of changes during the adaptation process is also critical, as opposed to only comparing the end states. It should be noted that motor adaptation operates across distinct timescales depending on the specific variable used to define adaptation. In addition, recent works suggest that complete convergence often requires extended or multi-session exposure. The observed kinematic convergence within the 25-minute walking employed in this study represents an early transient phase of the motor adaptation, rather than a full steady-state long-term adaptation. Similar to split-belt treadmill walking, motor adaptation in exoskeleton-assisted locomotion involves continuous motor variability reduction over time \cite{sanchez}. Unlike in No-Exo and Exo Zero-Torque tasks, where the metabolic responses quickly reached steady-state within the first few minutes of the experiment, the MCoT responses often did not converge to a stable steady-state during Exo-Assisted walking. While the complex physiological processes governing the metabolic response make it difficult to isolate the exact reasons for such divergence, factors such as fatigue from prolonged experiment, intricate dynamics in pHEI, and the unique neuromotor adaptation strategies would affect the MCoT response characteristics.  

In many exoskeleton studies, the steady-state metabolic response is often adopted as a primary indicator for perceived benefits of exoskeleton use \cite{katjaReviewPaper}. Not only it is difficult to pinpoint a reasonable homogeneous timing to declare steady-state due to the intricate nature of metabolic response, relying on an arbitrary duration for adopting steady-state assumption can also lead to erroneous interpretation. For example, P4 initially exhibited a descending trend in MCoT in the first 12 minutes, followed by a period of apparent steady-state that later deviated toward increased MCoT. Similarly, P1 demonstrated over 20\% of increased MCoT during the first 10 minutes of the study, which then reduced to less than 10\% of increased MCoT. In this both cases, the interpretation on the benefit of the exoskeleton vary depending on at which minute the steady-state assumption was adopted. 



Even though the subjective perception aspect was limited by sample size, it provided an intriguing perspective toward how humans conceptualize adaptation during exoskeleton-assisted locomotion. For instance, the joint-level inter-trajectory RMSE revealed that all participants achieved kinematic convergence at the ankle and hip joints, as well as the knee joint for 7 out of 9 participants. However, only 5 out of 9 participants self-reported feeling adapted to exoskeleton use. Notably, all 5 of these participants exhibited decreasing trend in their inter-trajectory RMSE while reporting adaptation. Although the multivariate nature of human adaptation behavior makes it challenging to pinpoint the exact mechanisms behind this mismatch, these findings suggest that the subjective perception of adaptation is not purely driven by achieving kinematic steady-state. Instead, psychological factors such as familiarization and trust in the device likely play a major role, emphasizing the need to incorporate the psychological metrics to bridge the gap between biomechanical convergence and psychological exoskeleton familiarization.

The choice of temporal discretization can also influence the interpretation of kinematic adaptation results. In this study, the joint kinematics were discretized using a static 15-second averaging window to balance between robustness against measurement noise and the preservation of the rapid dynamics during the earliest phase of the short-term motor adaptation. Averaging across excessively short windows increases sensitivity to stride-to-stride variability and amplifies noise, whereas overly long windows risk obscuring transient adaptation behavior. When the data was discretized using windows shorter than 15 seconds, the results became dominated by the variance in the data, making trend identification unreliable, particularly during Exo-Assisted walking. This reflects a methodological compromise between noise suppression and preservation of meaningful adaptation dynamics, in which alternative window choices may influence results interpretation.
\section{Conclusion}
This study demonstrates that motor adaptation to exoskeleton-assisted locomotion is a gradual, individualized process, with substantial inter-participant variability in both the timing and magnitude of biomechanical variables such as kinematics and metabolic response. While most participants eventually achieved convergence in joint-level trajectories using the proposed quantitative metric for gauging kinematic convergence, the asynchrony in adaptation across lower limb joints underscored the complexity of adaptation processes in physical human-exoskeleton interaction. The findings from this work also highlighted that group-averaged analyses can obscure meaningful inter-participant differences, potentially leading to overestimation of exoskeleton-assisted performance or adaptation.

Objective metrics, such as joint-level inter-trajectory RMSE and linear dependency analysis, provided complementary insights regarding the natural adaptation dynamics, whereas the self-reported subjective perception of adaptation often did not correspond to the measured changes in biomechanical variables. These results emphasized the need for better definition of objective and perceived adaptation, and the need for combined subjective and objective assessment strategies when evaluating exoskeleton-assisted locomotion.

Finally, methodological considerations such as kinematics-based gait segmentation, choice of temporal discretization window of kinematic data, and rationale behind the adopted strategies were shared throughout. Overall, this work provided a framework for assessing individualized motor adaptation to exoskeleton assistance and emphasized the importance of accounting for variability, both within and across participants, in the evaluation of lower limb exoskeleton assistance. 


\bibliographystyle{IEEEtran}
\bibliography{./ref}

@INPROCEEDINGS{giorgos,
    author={Marinou, Giorgos and Sloot, Lizeth and Mombaur, Katja},
    booktitle={2022 9th IEEE RAS/EMBS International Conference for Biomedical Robotics and Biomechatronics (BioRob)}, 
    title={Towards efficient lower-limb exoskeleton evaluation: defining biomechanical metrics to quantify assisted gait familiarization}, 
    year={2022},
    pages={1-8},
    doi={10.1109/BioRob52689.2022.9925360}}

@article{Kirchner,
    author = {Kirchner, Elsa Andrea and B{\"u}tef{\"u}r, Judith},
    title = {Towards Bidirectional and Coadaptive Robotic Exoskeletons for Neuromotor Rehabilitation and Assisted Daily Living: a Review},
    journal = {Current Robotics Reports},
    year = {2022},
    month = {Jun},
    day = {1},
    volume = {3},
    number = {2},
    pages = {21-32},
    doi = {10.1007/s43154-022-00076-7}}

@article{h2tPaper,
  author={Dežman, Miha and Marquardt, Charlotte and Üğür, Adnan and Moeller, Tobias and Asfour, Tamim},
  journal={IEEE Transactions on Medical Robotics and Bionics}, 
  title={Influence of Motion Restrictions in an Ankle Exoskeleton on Gait Kinematics and Stability in Straight Walking}, 
  year={2025},
  volume={7},
  number={1},
  pages={114-122},
  doi={10.1109/TMRB.2024.3503896}}

@article{katjaReviewPaper,
  author={Pinto-Fernandez, David and Torricelli, Diego and Sanchez-Villamanan, Maria del Carmen and Aller, Felix and Mombaur, Katja and Conti, Roberto and Vitiello, Nicola and Moreno, Juan C. and Pons, Jose Luis},
  journal={IEEE Transactions on Neural Systems and Rehabilitation Engineering}, 
  title={Performance Evaluation of Lower Limb Exoskeletons: A Systematic Review}, 
  year={2020},
  volume={28},
  number={7},
  pages={1573-1583},
  doi={10.1109/TNSRE.2020.2989481}}

@ARTICLE{conorWalshSoftExosuit,
  author={Ding, Ye and Galiana, Ignacio and Asbeck, Alan T. and De Rossi, Stefano Marco Maria and Bae, Jaehyun and Santos, Thiago Ribeiro Teles and de Araujo, Vanessa Lara and Lee, Sangjun and Holt, Kenneth G. and Walsh, Conor},
  journal={IEEE Transactions on Neural Systems and Rehabilitation Engineering}, 
  title={Biomechanical and Physiological Evaluation of Multi-Joint Assistance With Soft Exosuits}, 
  year={2017},
  volume={25},
  number={2},
  pages={119-130},
  doi={10.1109/TNSRE.2016.2523250}}

@article{avgParticipants,
  title = {An impairment-specific hip exoskeleton assistance for gait training in subjects with acquired brain injury: a feasibility study},  
  author = {Livolsi, Chiara and Conti, Roberto and Guanziroli, Eleonora and Fridriksson, Thor and Alexandersson, Asgeir and Kristjansson, Kristleifur and Esquenazi, Alberto and Molino Lova, Raffaele and Romo, Duane and Giovacchini, Francesco and Crea, Simona and Molteni, Franco and Vitiello, Nicola},
  journal = {Scientific Reports},
  year = {2022},
  month = nov,
  day={11},
  volume={12},
  number={19343},
  DOI={10.1038/s41598-022-23283-w}}

@article{interSubject,
  author = {Lora-Millan, Julio Salvador and Sanchez-Cuesta, Francisco José and Romero, Juan Pablo and Moreno, Juan C. and Rocon, Eduardo},  
  title = {Robotic exoskeleton embodiment in post-stroke hemiparetic patients: an experimental study about the integration of the assistance provided by the REFLEX knee exoskeleton},
  journal = {Scientific Reports},  
  year={2023},  
  month = dec,
  day={21},
  volume = {13},
  DOI = {10.1038/s41598-023-50387-8},
  number = {22908}}

@article{25min,
    title = {Adaptation to walking with an exoskeleton that assists ankle extension},
    author = {S. Galle and P. Malcolm and W. Derave and D. {De Clercq}},    
    journal = {Gait \& Posture},
    volume = {38},
    number = {3},
    pages = {495-499},
    year = {2013},
    doi = {https://doi.org/10.1016/j.gaitpost.2013.01.029}}

@ARTICLE{2min,
  author={Zhang, Xiaohui and Tricomi, Enrica and Missiroli, Francesco and Lotti, Nicola and Bokranz, Casimir and Nicklas, Daniela and Masia, Lorenzo},
  journal={IEEE Robotics and Automation Letters}, 
  title={Enhancing Gait Assistance Control Robustness of a Hip Exosuit by Means of Machine Learning}, 
  year={2022},
  volume={7},
  number={3},
  pages={7566-7573},
  doi={10.1109/LRA.2022.3183791}}

@article{sparkOG,
  author = {Orekhov, Greg and Fang, Ying and Cuddeback, Chance F. and Lerner, Zachary F.},
  journal = {Journal of NeuroEngineering and Rehabilitation},
  title = {Usability and performance validation of an ultra-lightweight and versatile untethered robotic ankle exoskeleton},
  year = {2021},
  volume = {18},
  number = {163},
  DOI = {10.1186/s12984-021-00954-9},
  month = nov,
  day={10}}

@article{spark,
    author={Gasparri, Gian Maria and Luque, Jason and Lerner, Zachary F.},
    journal={IEEE Transactions on Neural Systems and Rehabilitation Engineering}, 
    title={Proportional Joint-Moment Control for Instantaneously Adaptive Ankle Exoskeleton Assistance}, 
    year={2019},
    volume={27},
    number={4},
    pages={751-759},
    doi={10.1109/TNSRE.2019.2905979}}

@article{MET,
    author = {B. E. Ainsworth and W. L. Haskell and M. C. Whitt and M. L. Irwin and A. M. Swartz and S. J. Strath and W. L. O'Brien and D. R. Bassett Jr. and K. H. Schmitz and P. O. Emplaincourt and D. R. Jacobs Jr. and A. S. Leon},
    title = {Compendium of physical activities: an update of activity codes and MET intensities},
    journal = {Medicine \& Science in Sports \& Exercise},
    volume = {32},
    number = {9},
    pages = {498-516},
    year = {2000}}

@article{metabolicNoFood,
  title = {Best Practice Methods to Apply to Measurement of Resting Metabolic Rate in Adults: A Systematic Review},
  author = {Compher, Charlene and Frankenfield, David and Keim, Nancy and Roth-Yousey, Lori},
  year = {2006},
  month = jun,
  journal = {Journal of the American Dietetic Association},
  volume = {106},
  number = {6},
  pages = {881--903},
  doi = {10.1016/j.jada.2006.02.009}}

@article{6minTreadmill,
  title = {Familiarization with treadmill walking: How much is enough?},
  volume = {9},
  DOI = {10.1038/s41598-019-41721-0},
  number = {5232},
  journal = {Scientific Reports},
  publisher = {Springer Science and Business Media LLC},
  author = {Meyer,  Christian and Killeen,  Tim and Easthope,  Christopher S. and Curt,  Armin and Bolliger,  Marc and Linnebank,  Michael and Z\"{o}rner,  Bj\"{o}rn and Filli,  Linard},
  year = {2019},
  month = mar }

@article{25min_2,
author = {Keith E. Gordon and Daniel P. Ferris},
title = {Learning to walk with a robotic ankle exoskeleton},
journal = {Journal of Biomechanics},
volume = {40},
number = {12},
pages = {2636-2644},
year = {2007},
doi = {https://doi.org/10.1016/j.jbiomech.2006.12.006}}

@article{toeOffDetection,
  title = {Two simple methods for determining gait events during treadmill and overground walking using kinematic data},
  volume = {27},
  DOI = {10.1016/j.gaitpost.2007.07.007},
  number = {4},
  journal = {Gait \& amp; Posture},
  publisher = {Elsevier BV},
  author = {Zeni,  J.A. and Richards,  J.G. and Higginson,  J.S.},
  year = {2008},
  month = may,
  pages = {710–714}}

@article{heelStrikeDetection,
title = {A marker based kinematic method of identifying initial contact during gait suitable for use in real-time visual feedback applications},
journal = {Gait \& Posture},
volume = {36},
number = {3},
pages = {650-652},
year = {2012},
issn = {0966-6362},
doi = {https://doi.org/10.1016/j.gaitpost.2012.04.016},
author = {A.R. {De Asha} and M.A. Robinson and G.J. Barton}}

@article{brockwayEquation,
  title={Derivation of formulae used to calculate energy expenditure in man.},
  author={J. M. Brockway},
  journal={Human nutrition. Clinical nutrition},
  year={1987},
  volume={41},
  number={6},
  pages={463-471}}

@article{speed,
    author = {Juanjuan Zhang  and Pieter Fiers  and Kirby A. Witte  and Rachel W. Jackson  and Katherine L. Poggensee  and Christopher G. Atkeson  and Steven H. Collins },
    title = {Human-in-the-loop optimization of exoskeleton assistance during walking},
    journal = {Science Current Robotics Reports},
    volume = {356},
    number = {6344},
    pages = {1280-1284},
    year = {2017},
    doi = {10.1126/science.aal5054}}

@article{subconsciousAdaptation,
  author = {Wilkenfeld, J Nan and Kim, Sunwook and Upasani, Satyajit and Kirkwood, Gavin and Dunbar, Norah and Srinivasan, Divya},
  journal = {Frontiers in Robotics and AI},
  title = {Sensemaking, adaptation and agency in human-exoskeleton synchrony},
  year = {2023},
  volume = {10},
  DOI = {10.3389/frobt.2023.1207052},
  month = oct,
  day={12}}

@article{theiaValidLowerLimbSagittal,
title = {Construct validity of markerless three-dimensional gait biomechanics in healthy older adults},
journal = {Gait \& Posture},
volume = {120},
pages = {217-225},
year = {2025},
issn = {0966-6362},
doi = {https://doi.org/10.1016/j.gaitpost.2025.04.022},
author = {Andreia Carvalho and Jos Vanrenterghem and Sílvia Cabral and Ana {M. d'Assunção} and Filomena Carnide and António P. Veloso and Vera Moniz-Pereira}}

@article{theiaValidWalkingRunning,
author = {Josh Walker and Aaron Thomas and David E. Lunn and Gareth Nicholson and Catherine B. Tucker},
title = {Concurrent validity of Theia3D markerless motion capture for detecting sagittal kinematic differences between gait speeds},
journal = {Journal of Sports Sciences},
volume = {43},
number = {16},
pages = {1560--1571},
year = {2025},
doi = {10.1080/02640414.2025.2513151}}

@article{exoReviewClinical,
  title = {Systematic Review on Wearable Lower-Limb Exoskeletons for Gait Training in Neuromuscular Impairments},
  author = {{Rodr{\'i}guez-Fern{\'a}ndez}, Antonio and {Lobo-Prat}, Joan and {Font-Llagunes}, Josep M.},
  year = 2021,
  month = dec,
  journal = {Journal of NeuroEngineering and Rehabilitation},
  volume = {18},
  number = {1},
  pages = {22},
  issn = {1743-0003},
}

@Article{exoReviewIntention,
    AUTHOR = {Wang, Tao and Zhang, Bin and Liu, Chenhao and Liu, Tao and Han, Yi and Wang, Shuoyu and Ferreira, João P. and Dong, Wei and Zhang, Xiufeng},
    TITLE = {A Review on the Rehabilitation Exoskeletons for the Lower Limbs of the Elderly and the Disabled},
    JOURNAL = {Electronics},
    VOLUME = {11},
    YEAR = {2022},
    NUMBER = {3},
    ARTICLE-NUMBER = {388},
    URL = {https://www.mdpi.com/2079-9292/11/3/388},
    ISSN = {2079-9292},
    DOI = {10.3390/electronics11030388}
}

@Article{exoReviewPHRI,
    AUTHOR = {Massardi, Stefano and Rodriguez-Cianca, David and Pinto-Fernandez, David and Moreno, Juan C. and Lancini, Matteo and Torricelli, Diego},
    TITLE = {Characterization and Evaluation of Human–Exoskeleton Interaction Dynamics: A Review},
    JOURNAL = {Sensors},
    VOLUME = {22},
    YEAR = {2022},
    NUMBER = {11},
    ARTICLE-NUMBER = {3993},
    URL = {https://www.mdpi.com/1424-8220/22/11/3993},
    PubMedID = {35684614},
    ISSN = {1424-8220},
    DOI = {10.3390/s22113993}
}

@ARTICLE{pHEIComplexity,
  author={Chen, Longbao and Zhou, Ding and Leng, Yuquan},
  journal={IEEE Transactions on Neural Systems and Rehabilitation Engineering}, 
  title={A Systematic Review on Rigid Exoskeleton Robot Design for Wearing Comfort: Joint Self-Alignment, Attachment Interface, and Structure Customization}, 
  year={2024},
  volume={32},
  number={},
  pages={3815-3827},
  doi={10.1109/TNSRE.2024.3479283}
}

@Article{pHEIIssues,
AUTHOR = {Moeller, Tobias and Moehler, Felix and Krell-Roesch, Janina and Dežman, Miha and Marquardt, Charlotte and Asfour, Tamim and Stein, Thorsten and Woll, Alexander},
TITLE = {Use of Lower Limb Exoskeletons as an Assessment Tool for Human Motor Performance: A Systematic Review},
JOURNAL = {Sensors},
VOLUME = {23},
YEAR = {2023},
NUMBER = {6},
ARTICLE-NUMBER = {3032},
PubMedID = {36991743},
ISSN = {1424-8220},
DOI = {10.3390/s23063032}
}

@article{pHEIPortrait,
  title = {Human--Exoskeleton Interaction Portrait},
  author = {Shushtari, Mohammad and Foellmer, Julia and Arami, Arash},
  year = 2024,
  month = sep,
  journal = {Journal of NeuroEngineering and Rehabilitation},
  volume = {21},
  number = {1},
  pages = {152},
  issn = {1743-0003},
  doi = {10.1186/s12984-024-01447-1},
}

@article{individualDiff,
  title = {Reducing the Metabolic Cost of Walking with an Ankle Exoskeleton: Interaction between Actuation Timing and Power},
  shorttitle = {Reducing the Metabolic Cost of Walking with an Ankle Exoskeleton},
  author = {Galle, Samuel and Malcolm, Philippe and Collins, Steven Hartley and De Clercq, Dirk},
  year = 2017,
  month = dec,
  journal = {Journal of NeuroEngineering and Rehabilitation},
  volume = {14},
  number = {1},
  pages = {35},
  issn = {1743-0003},
  doi = {10.1186/s12984-017-0235-0},
}

@article{conorWalshMetabolic,
  title = {Metabolic Cost Adaptations during Training with a Soft Exosuit Assisting the Hip Joint},
  author = {Panizzolo, Fausto A. and Freisinger, Gregory M. and Karavas, Nikos and {Eckert-Erdheim}, Asa M. and Siviy, Christopher and Long, Andrew and Zifchock, Rebecca A. and LaFiandra, Michael E. and Walsh, Conor J.},
  year = 2019,
  month = jul,
  journal = {Scientific Reports},
  volume = {9},
  number = {1},
  pages = {9779},
  issn = {2045-2322},
  doi = {10.1038/s41598-019-45914-5},
}

@article{sparkmore,
author = {Benjamin C. Conner and Alyssa M. Spomer and Katherine M. Steele and Zachary F. Lerner},
title = {Factors influencing neuromuscular responses to gait training with a robotic ankle exoskeleton in cerebral palsy},
journal = {Assistive Technology},
volume = {35},
number = {6},
pages = {463--470},
year = {2023},
publisher = {Taylor \& Francis},
doi = {10.1080/10400435.2022.2121324},
}

@article{sanchez,
  title = {Evidence of {{Energetic Optimization}} during {{Adaptation Differs}} for {{Metabolic}}, {{Mechanical}}, and {{Perceptual Estimates}} of {{Energetic Cost}}},
  author = {Sánchez, Natalia and Park, Sungwoo and Finley, James M.},
  date = {2017-08-09},
  year={2017},
  journal = {Scientific Reports},
  shortjournal = {Sci Rep},
  volume = {7},
  number = {1},
  pages = {7682},
  issn = {2045-2322},
  doi = {10.1038/s41598-017-08147-y}
}

@article{poggensee2021,
  title = {How Adaptation, Training, and Customization Contribute to Benefits from Exoskeleton Assistance},
  author = {Poggensee, Katherine L. and Collins, Steven H.},
  date = {2021-09-29},
  year={2021},
  journal = {Science Robotics},
  shortjournal = {Sci. Robot.},
  volume = {6},
  number = {58},
  pages = {eabf1078},
  issn = {2470-9476},
  doi = {10.1126/scirobotics.abf1078}
}

@article{panizzolo2019,
  title = {Metabolic Cost Adaptations during Training with a Soft Exosuit Assisting the Hip Joint},
  author = {Panizzolo, Fausto A. and Freisinger, Gregory M. and Karavas, Nikos and Eckert-Erdheim, Asa M. and Siviy, Christopher and Long, Andrew and Zifchock, Rebecca A. and LaFiandra, Michael E. and Walsh, Conor J.},
  date = {2019-07-05},
  year={2019},
  journal = {Scientific Reports},
  shortjournal = {Sci Rep},
  volume = {9},
  number = {1},
  pages = {9779},
  issn = {2045-2322},
  doi = {10.1038/s41598-019-45914-5}
}

@article{hybart2023,
  title = {Neuromechanical {{Adaptation}} to {{Walking With Electromechanical Ankle Exoskeletons Under Proportional Myoelectric Control}}},
  author = {Hybart, Rachel L. and Ferris, Daniel P.},
  journal = {IEEE Open Journal of Engineering in Medicine and Biology},
  year={2023},
  shortjournal = {IEEE Open J. Eng. Med. Biol.},
  volume = {4},
  pages = {119--128},
  issn = {2644-1276},
  doi = {10.1109/OJEMB.2023.3288469}
}

@article{abram2022,
  title = {General Variability Leads to Specific Adaptation toward Optimal Movement Policies},
  author = {Abram, Sabrina J. and Poggensee, Katherine L. and Sánchez, Natalia and Simha, Surabhi N. and Finley, James M. and Collins, Steven H. and Donelan, J. Maxwell},
  date = {2022-05},
  year={2022},
  journal = {Current Biology},
  shortjournal = {Current Biology},
  volume = {32},
  number = {10},
  pages = {2222-2232.e5},
  issn = {09609822},
  doi = {10.1016/j.cub.2022.04.015}
}

@article{Orehhov2020,
  title = {Ankle {{Exoskeleton Assistance Can Improve Over-Ground Walking Economy}} in {{Individuals With Cerebral Palsy}}},
  author = {Orekhov, Greg and Fang, Ying and Luque, Jason and Lerner, Zachary F.},
  date = {2020-02},
  year={2020},
  journal = {IEEE Transactions on Neural Systems and Rehabilitation Engineering},
  shortjournal = {IEEE Trans. Neural Syst. Rehabil. Eng.},
  volume = {28},
  number = {2},
  pages = {461--467},
  issn = {1534-4320, 1558-0210},
  doi = {10.1109/TNSRE.2020.2965029}
}

@article{fang2022,
  title = {Adaptive Ankle Exoskeleton Gait Training Demonstrates Acute Neuromuscular and Spatiotemporal Benefits for Individuals with Cerebral Palsy: {{A}} Pilot Study},
  shorttitle = {Adaptive Ankle Exoskeleton Gait Training Demonstrates Acute Neuromuscular and Spatiotemporal Benefits for Individuals with Cerebral Palsy},
  author = {Fang, Ying and Orekhov, Greg and Lerner, Zachary F.},
  date = {2022-06},
  year={2022},
  journal = {Gait \& Posture},
  shortjournal = {Gait \& Posture},
  volume = {95},
  pages = {256--263},
  issn = {09666362},
  doi = {10.1016/j.gaitpost.2020.11.005}
}

@article{hicheur2006,
  title = {Intersegmental {{Coordination During Human Locomotion}}: {{Does Planar Covariation}} of {{Elevation Angles Reflect Central Constraints}}?},
  shorttitle = {Intersegmental {{Coordination During Human Locomotion}}},
  author = {Hicheur, Halim and Terekhov, Alexander V. and Berthoz, Alain},
  date = {2006-09},
  year={2006},
  journal = {Journal of Neurophysiology},
  shortjournal = {Journal of Neurophysiology},
  volume = {96},
  number = {3},
  pages = {1406--1419},
  issn = {0022-3077, 1522-1598},
  doi = {10.1152/jn.00289.2006}
}

@article{lora-millan2022,
  title = {Coordination {{Between Partial Robotic Exoskeletons}} and {{Human Gait}}: {{A Comprehensive Review}} on {{Control Strategies}}},
  shorttitle = {Coordination {{Between Partial Robotic Exoskeletons}} and {{Human Gait}}},
  author = {Lora-Millan, Julio S. and Moreno, Juan C. and Rocon, E.},
  date = {2022-05-25},
  year={2022},
  journal = {Frontiers in Bioengineering and Biotechnology},
  shortjournal = {Front. Bioeng. Biotechnol.},
  volume = {10},
  pages = {842294},
  issn = {2296-4185},
  doi = {10.3389/fbioe.2022.842294}
}

@article{lee2024,
  title = {The {{Effectiveness}} of {{Overground Robot Exoskeleton Gait Training}} on {{Gait Outcomes}}, {{Balance}}, and {{Motor Function}} in {{Patients}} with {{Stroke}}: {{A Systematic Review}} and {{Meta-Analysis}} of {{Randomized Controlled Trials}}},
  shorttitle = {The {{Effectiveness}} of {{Overground Robot Exoskeleton Gait Training}} on {{Gait Outcomes}}, {{Balance}}, and {{Motor Function}} in {{Patients}} with {{Stroke}}},
  author = {Lee, Myoung-Ho and Tian, Ming-Yu and Kim, Myoung-Kwon},
  date = {2024-08-19},
  year={2024},
  journal = {Brain Sciences},
  shortjournal = {Brain Sciences},
  volume = {14},
  number = {8},
  pages = {834},
  issn = {2076-3425},
  doi = {10.3390/brainsci14080834}
}

\end{document}